\documentclass{article}
\pdfoutput=1

\usepackage{natbib}
\setcitestyle{numbers,square,comma,sort}

\usepackage[preprint]{neurips_2024}

\usepackage[utf8]{inputenc} 
\usepackage[T1]{fontenc}    
\usepackage{url}            
\usepackage{booktabs}       
\usepackage{amsfonts}       
\usepackage{nicefrac}       
\usepackage{microtype}      
\usepackage{colortbl}
\usepackage[dvipsnames]{xcolor}

\usepackage{graphicx}
\usepackage{amsmath}
\usepackage{amssymb}
\usepackage{booktabs}
\usepackage[normalem]{ulem} 

\usepackage{algorithmic}
\usepackage{algorithm}

\usepackage{multirow}
\usepackage{caption}
\usepackage{subcaption}

\usepackage{wrapfig}
\usepackage[figuresright]{rotating}

\definecolor{robo_blue}{RGB}{99, 113, 250}
\definecolor{robo_red}{RGB}{239, 99, 75}
\definecolor{robo_green}{RGB}{0, 180, 139}

\usepackage[pagebackref=true,colorlinks,urlcolor=brown,citecolor=brown,linkcolor=brown]{hyperref}
\usepackage{cleveref}

\title{PoseBench: Benchmarking the Robustness of Pose Estimation Models under Corruptions}
\author{%
  Sihan Ma\textsuperscript{1},
  Jing Zhang\textsuperscript{1},
  Qiong Cao\textsuperscript{2},
  Dacheng Tao\textsuperscript{3}
  \\
   \textsuperscript{1}The University of Sydney, Australia \
   \textsuperscript{2}JD Explore Academy, China \\
   \textsuperscript{3}Nanyang Technological University, Singapore
  \\
  \texttt{sima7436@uni.sydney.edu.au}\\
  \texttt{\{jingzhang.cv, mathqiong2012, dacheng.tao\}@gmail.com} \\
  \url{https://xymsh.github.io/PoseBench/}
}

\begin{document}

\maketitle

\begin{abstract}
Pose estimation aims to accurately identify anatomical keypoints in humans and animals using monocular images, which is crucial for various applications such as human-machine interaction, embodied AI, and autonomous driving. While current models show promising results, they are typically trained and tested on clean data, potentially overlooking the corruption during real-world deployment and thus posing safety risks in practical scenarios. To address this issue, we introduce PoseBench, a comprehensive benchmark designed to evaluate the robustness of pose estimation models against real-world corruption. We evaluated 60 representative models, including top-down, bottom-up, heatmap-based, regression-based, and classification-based methods, across three datasets for human and animal pose estimation. Our evaluation involves 10 types of corruption in four categories: 1) blur and noise, 2) compression and color loss, 3) severe lighting, and 4) masks. Our findings reveal that state-of-the-art models are vulnerable to common real-world corruptions and exhibit distinct behaviors when tackling human and animal pose estimation tasks. To improve model robustness, we delve into various design considerations, including input resolution, pre-training datasets, backbone capacity, post-processing, and data augmentations. We hope that our benchmark will serve as a foundation for advancing research in robust pose estimation. The benchmark and source code will be released at \href{https://xymsh.github.io/PoseBench/}{PoseBench}.
\end{abstract}

\section{Introduction}\label{sec:intro}

Pose estimation aims to locate anatomical keypoints of the human or animal body from a single image~\cite{survey2023, survey2021}. This fundamental ability to accurately interpret human movements drives advancements across various fields, enabling innovative applications in healthcare, entertainment, safety, autonomous driving, and wildlife conservation~\cite{iotsurvey2020, pavlakos2018learning, lamas2022human, kumar2021omnidet, hahn2021deep}. 
Existing pose estimation models achieve remarkable performance on various benchmarks~\cite{coco, mpii, ochuman, crowdpose, ap10k} through two typical paradigms: top-down~\cite{hourglass, simplebaseline, hrnet, zhang2021towards, hrformer, vitpose, xu2023vitpose++} and bottom-up~\cite{ae, higherhrnet, dekr, xu2023vitpose++}. These paradigms differ in whether they detect individuals at the beginning or end of the process. Based on the keypoint prediction heads, top-down methods can further be categorized into approaches using 2D Gaussian heatmaps centered at keypoints~\cite{hrnet, mspn, rsn, transpose, vitpose}, regression of keypoint locations~\cite{prtr, poseur}, or classification of keypoints~\cite{simcc}. All the paradigms have their unique advantages, \textit{e.g.}, in terms of accuracy or running speed, contributing to the versatility and effectiveness of pose estimation in diverse applications.

Despite significant advancements in pose estimation, existing models often overlook the challenges of real-world deployment. Different from the ``clean'' images used during training, images from cameras deployed in real-world scenarios can be corrupted at various stages - during sensing, signal transmission, or data storage - introducing natural corruptions such as motion blur, noise, low lighting, and compression artifacts~\cite{roboposebenchmark, robotad}. These issues extend beyond the scope of current pose datasets, making pose models trained on them vulnerable to such corruptions and raising safety concerns in practical applications. 

To address this issue, it is necessary to establish a comprehensive benchmark to assess the robustness of current pose estimation methods and to devise an effective strategy against real-world corruptions. Recent studies have explored robustness benchmarks for 3D perception tasks in autonomous driving, such as depth estimation~\cite{robodepth}, 3D detection and segmentation~\cite{robo3d}. However, the robustness of pose estimation, a crucial location-sensitive task, remains under-explored. In 2021, \cite{roboposebenchmark} tested several pose estimation methods on common corruptions. This study, however, included only four convolutional neural networks (CNN)-based methods and did not adequately address broader pose estimation tasks including those for animals, which is insufficient to fully validate the robustness of current techniques, especially considering the many recent advancements in vision transformers (ViT)~\cite{dosovitskiy2020image,liu2021swin, xu2021vitae,zhang2023vitaev2,zhang2024vision,vitpose,xu2023vitpose++,simcc,hrformer,transpose}.

To fill this gap, we propose a more comprehensive robustness benchmark to evaluate existing representative CNN-based and ViT-based pose estimation models under a range of natural corruptions. Our benchmark assesses 60 models across various pose estimation methods, including top-down, bottom-up, heatmap-based, regression-based, and classification-based approaches, on both human and animal pose estimation tasks. We introduce four sets of corruptions encompassing 10 typical natural corruptions observed in real-world scenarios. To ensure a thorough evaluation, we consider five levels of severity for each type of corruption and report the average performance across all severity levels. Furthermore, we delve deeper into the impact of various configurations on robustness,
including input resolution, pre-training datasets, backbone capacity, post-processing, and data augmentations. This comprehensive evaluation enables a clear understanding of how different factors influence the robustness of pose estimation models under real-world conditions.

Extensive experiments were conducted to evaluate robustness, yielding several valuable findings:
1) Existing pose estimation models demonstrate a notable vulnerability to corruptions, yet their robustness is strongly correlated with performance on images without corruptions.
2) Among all the corruptions, motion blur and contrast have the most significant impact on model robustness, while brightness has the least impact.
3) Regression-based methods demonstrate the highest resistance to mask corruption, even though their performance on clean images is not the best.
4) The robustness against corruptions varies across existing datasets due to the different characteristics of the data. Human pose models are particularly vulnerable to compression and blur, whereas animal pose models experience the largest performance drop with contrast changes.
5) Among the various design factors, pre-training, post-processing, and large transformer-based backbone designs significantly enhance robustness, whereas input resolution does not have a noticeable effect.

The key contributions of this work are summarized as follows:
\begin{itemize}
    \item We introduce PoseBench, a comprehensive benchmark for evaluating the robustness of pose estimation models against various natural corruptions.
    \item We extensively evaluate 60 model variants from 15 state-of-the-art CNN-based and ViT-based pose estimation methods, covering top-down, bottom-up, heatmap-based, regression-based, and classification-based approaches. This includes testing their robustness against four sets of corruptions, featuring 10 types of common real-world corruptions.
    \item We investigate the impact of key design and training factors, such as backbone capacity, pre-training, input resolution, post-processing, and data augmentation. Our findings suggest effective strategies for enhancing both performance and robustness against corruptions.
\end{itemize}

\section{Related Work}\label{sec:related_work}

\subsection{Pose estimation}
Pose estimation aims to localize body joint positions from a single image of a human or animal and can be classified into two main approaches: top-down and bottom-up methods. 
Top-down methods~\cite{simplebaseline, hourglass, simplebaseline, hrnet, zhang2021towards, hrformer, transpose, rsn, mspn, vitpose, xu2023vitpose++, simcc} first detect individuals and then localize the keypoints within the bounding boxes of the detected individuals, offering greater accuracy and thus dominating the field. 
In contrast, bottom-up methods~\cite{ae,higherhrnet,dekr, xu2023vitpose++} first localize keypoints across the entire image, regardless of which individual they belong to, and then group these keypoints into distinct individuals. Bottom-up methods are more efficient and scalable for crowded scenes, although their accuracy heavily depends on the keypoint grouping technique. 

Within top-down methods, image features are extracted by the backbone architecture and then processed through the keypoint head for final prediction. Various methods, such as Hourglass~\cite{hourglass}, HRNet~\cite{hrnet}, MSPN~\cite{mspn}, RSN~\cite{rsn}, and SimCC~\cite{simcc}, utilize CNN-based backbones, including ResNet~\cite{resnet}, MobileNetV2~\cite{mobilenetv2}, and their variants. Recently, Vision Transformer (ViT)-based architectures, such as ViTPose~\cite{vitpose}, HRFormer~\cite{hrformer}, and Poseur~\cite{poseur}, have become widely used in pose estimation models, proving to be effective. Pose estimation approaches can be categorized into heatmap-based, regression-based, and classification-based methods. Heatmap-based methods~\cite{hourglass,vitpose,simcc,rsn,mspn,transpose,hrnet,hrformer,simplebaseline} generate 2D Gaussian distributions centered on the ground truth locations of keypoints, which becomes a standard practice in recent advancements. Classification-based methods~\cite{simcc} discretize coordinate locations to address quantization errors and computational upsampling processes inherent in heatmap methods. Regression-based methods~\cite{prtr,poseur} directly predict the coordinates of keypoints. 

Animal pose estimation~\cite{zhang2023clamp} has recently garnered significant attention with the emergence of datasets like AP10K~\cite{ap10k} and APT36K~\cite{apt36k}. Existing human pose estimation models can be readily adapted to these animal pose datasets, as the primary difference between the tasks is the skeleton structure. However, several challenges remain. In natural environments, high-speed animal movements and adverse weather conditions lead to significant motion blur, occlusions, and poor lighting, which greatly impact the robustness of pose estimation models.

\subsection{Robustness against corruptions}
Recent studies have explored robustness in various fields such as detection~\cite{robo3d}, segmentation~\cite{robo3d}, depth estimation~\cite{robodepth}, pose estimation~\cite{roboposebenchmark}, and action detection~\cite{robotad}, emphasizing models' generalization under corruptions due to safety concerns. Pose estimation robustness is particularly challenging because it involves both the classification of different human and animal species and pixel-level localization of keypoints. A previous seminal study~\cite{roboposebenchmark} evaluated the performance of four CNN-based models under corruptions like blur, noise, and lighting changes, providing valuable insights into robust pose estimation. Nevertheless, this study, conducted three years ago, did not include recent state-of-the-art models such as vision transformer-based methods (\textit{e.g.}, ViTPose~\cite{vitpose}, TransPose~\cite{transpose}, HRFormer~\cite{hrformer}) and classification-based methods (\textit{e.g.}, SimCC~\cite{simcc}). Moreover, it did not address animal pose estimation, which significantly differs from human pose estimation. To fill these gaps, we develop a more comprehensive benchmark to assess the robustness of 60 pose estimation models, covering 15 methods, against natural corruptions for both human and animal poses. We also delve into various factors for improving robustness, such as backbone capacity, pre-training, input resolution, post-processing, and data augmentation. Our findings offer interesting observations and valuable insights, which could inform the design of robust pose estimation models in the future.

\subsection{Data augmentation}
Data augmentation~\cite{autoaugment, cutout, rusak2020simple, cutmix, augphotometric, gridmask, augblur, augerase, geirhos2018generalisation} is a widely used and effective technique in computer vision tasks to enhance model generalization and robustness. By applying various transformations to the training data, it increases the dataset's size and diversity without the need to collect new data, making it easy to implement and highly effective. This process helps models learn more general features by exposing them to diverse data variations, thereby improving their ability to generalize to new, unseen data~\cite{robotad}. Common techniques include color changes, geometric transformations, blur and noise addition, and dropout. For instance, adding Gaussian or speckle noise~\cite{rusak2020simple} has been proven effective in improving generalization against various image corruptions. Research~\cite{augblur} has also highlighted the impact of blur on recognition tasks. Composite techniques that combine different transformations often yield better performance. AutoAugment~\cite{autoaugment}, for example, searches automatically for the optimal combination of augmentation policies. Additionally, methods like Cutmix~\cite{cutmix} and Mixup~\cite{mixup} enhance single images by merging them through overlapping, cutting, and pasting. In this study, we group different augmentations into four sets to assess their effectiveness in improving the robustness of existing pose estimation models. These sets include blur and noise, compression and color alteration, lighting adjustments, and occlusion with dropout.

\section{Benchmark}\label{sec:benchmark}
\begin{figure}[htb]
    \centering
    \includegraphics[width=\textwidth]{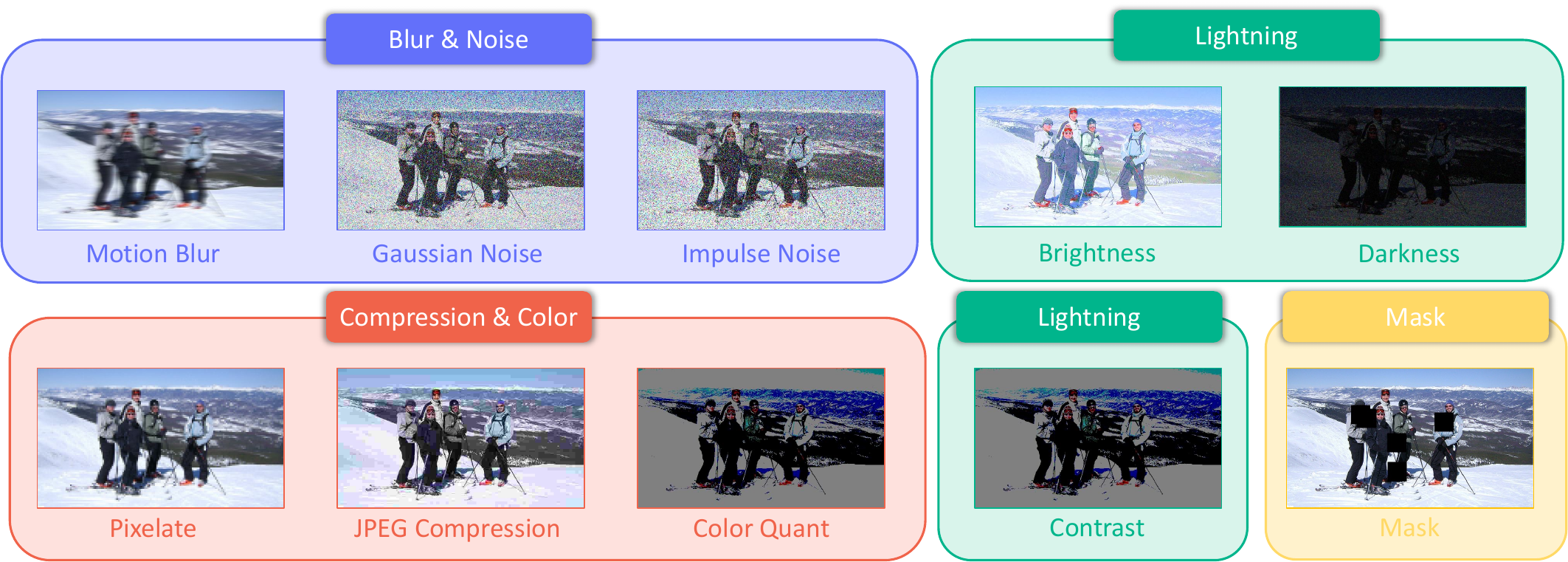} 
    \caption{Visualization of the four corruption types examined in this paper.}
    \label{fig:corruptions}
\end{figure}

In this paper, we investigate the robustness of 60 pose estimation models, categorized into top-down or bottom-up methods using heatmap-based, regression-based, or classification-based representations. We examine four groups of natural corruptions, covering 10 common types that could affect the robustness of pose estimation models in real-world scenarios, as illustrated in Figure ~\ref{fig:corruptions}. These corruptions include blur and noise, compression and color loss, lighting, and masks. To ensure comprehensive and fair evaluation, we evaluate human pose estimation models using the COCO~\cite{coco} and OCHuman~\cite{ochuman} datasets, and animal pose estimation models using the AP10K~\cite{ap10k} dataset, which are representative datasets in the field. Each corruption type is applied at five levels of severity to the validation sets, maintaining consistency across all methods. The results averaged over these five severity levels form a thorough comparison. This section details the types of corruptions (Section~\ref{sec:corr_types}) and the evaluation metrics used to measure robustness (Section~\ref{sec:metrics}).

\begin{wrapfigure}{r}{0.562\textwidth}
    \vspace{-0.56cm}
    \begin{center}
    \includegraphics[width=0.562\textwidth]{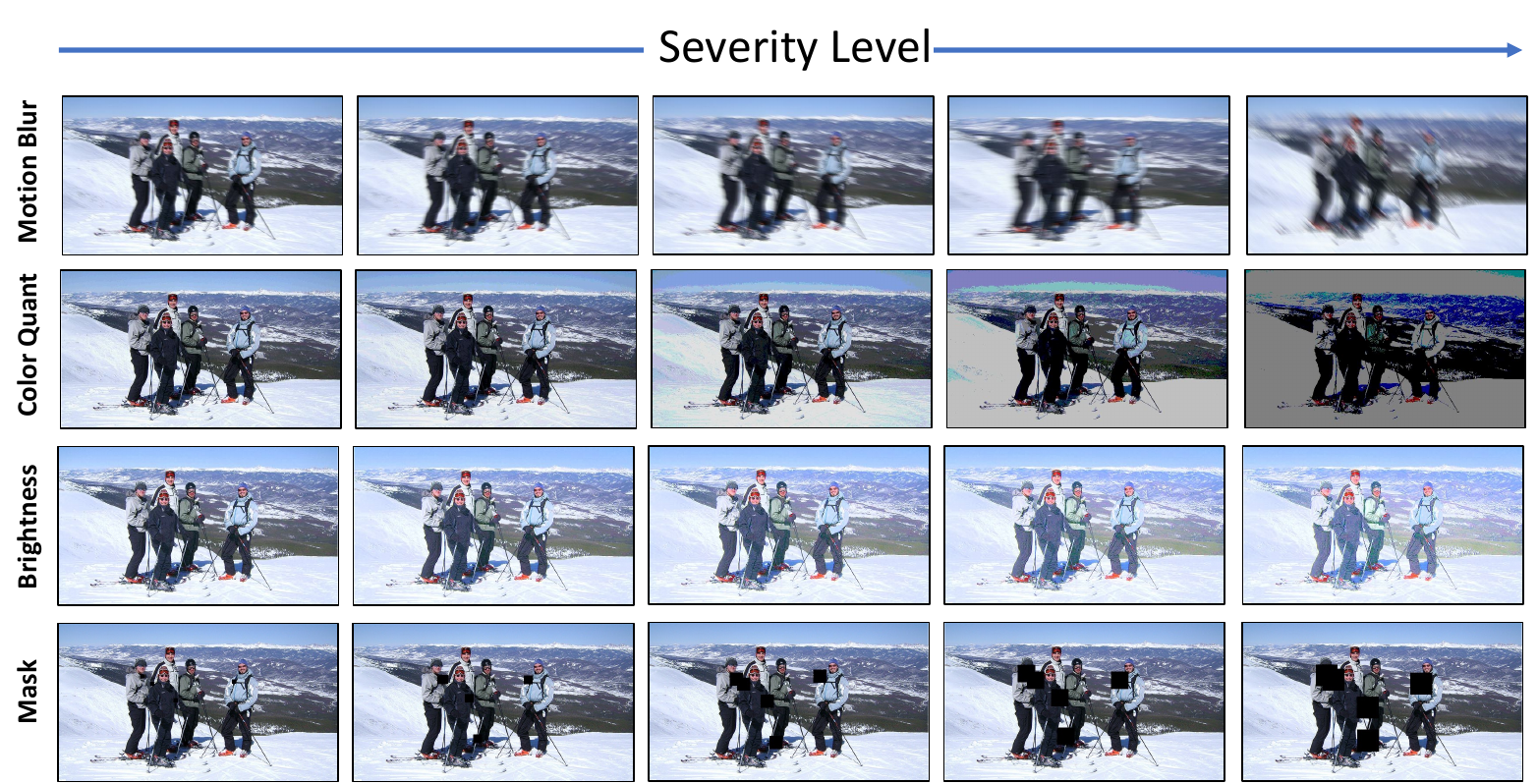}
    \end{center}
    \vspace{-0.2cm}
    \caption{Corruption severity level. Each corruption consists of five severity levels.}
    \label{fig:severity_level}
    \vspace{-0.3cm}
\end{wrapfigure}

\subsection{Corruptions}\label{sec:corr_types}
\paragraph{Blur \& Noise.} 
When deploying pose estimation models in real-world scenarios, it is crucial that they reliably handle issues arising during signal capture. Addressing these challenges significantly enhances system reliability and robustness, while reducing repair and maintenance costs. These issues include motion blur from object or sensor movement, noise from sensor hardware malfunctions, and noise from unstable signal transmission. To simulate image quality degradation under these conditions, we use three common types of corruption: Motion Blur, Gaussian Noise, and Impulse Noise. The visual effects of Motion Blur at varying severity levels are shown in the first row of Figure~\ref{fig:severity_level}.

\paragraph{Compression \& Color Loss.}
Image compression and color loss pose significant challenges for pose estimation models deployed in practical applications due to storage, transmission, and device limitations. These issues result in reduced color detail and fidelity, along with color banding and artifacts that impair pose estimation performance. To assess the impact, we simulate three types of compression and color-related corruptions: Pixelate, JPEG Compression, and Color Quant. The visual effects of Color Quant at varying severity levels are shown in the second row of Figure~\ref{fig:severity_level}.

\paragraph{Lightning.}
Lighting changes, such as extremely dark environments at night, excessive brightness from sunlight, and sensor malfunctions causing overexposure, are common issues in real-world scenarios. These conditions impair visibility and obscure essential features needed for accurate pose estimation. Current datasets lack representation of diverse lighting conditions, resulting in pose estimation models that are not robust under these circumstances. To address this, we simulate various lighting conditions using three types of corruptions: Brightness, Darkness, and Contrast. The visual effects of Brightness at different severity levels are illustrated in the third row of Figure~\ref{fig:severity_level}.

\paragraph{Mask.}
While occlusion has been extensively studied in pose estimation, current datasets mainly address occlusions caused by environments or other human bodies, which provide ample context for inferring hidden parts. However, in real-world scenarios, occlusions can also result from data loss during sensing, signal transmission, data processing, or privacy protection, where parts of the image are missing without any contextual clues. To simulate this, we apply random-size masks to keypoints to evaluate model robustness under mask corruption, as shown in Figure~\ref{fig:severity_level}. This method enables us to assess model performance when dealing with occlusions that lack contextual information.

\subsection{Benchmark Datasets}\label{sec:datasets}
Inspired by~\cite{robo3d,robodepth,roboposebenchmark}, our robustness benchmark for pose estimation comprises two tasks, utilizing three datasets termed COCO-C and OCHuman-C for human body pose estimation and AP10K-C for animal pose estimation. These datasets are created by applying four corruption sets introduced in Section~\ref{sec:corr_types}, covering 10 types of corruptions, to their original validation sets. Each corruption type includes five levels of severity.

\paragraph{COCO-C Dataset.}
The COCO-C dataset is constructed from the validation set of the COCO Keypoints 2017 dataset~\cite{coco}, which includes over 118K labeled images for training and 5000 images for validation. The COCO~\cite{coco} dataset supports a 17-keypoint annotation for human bodies, facilitating comprehensive human body pose estimation under various scenarios.

\paragraph{OCHuman-C Dataset.}
The OCHuman-C dataset is based on the OCHuman~\cite{ochuman} dataset, which contains 13,360 meticulously annotated human instances within 5,081 images. The OCHuman dataset is considered the most complex and challenging for human pose estimation due to the heavy occlusion of humans in this dataset.

\paragraph{AP10K-C Dataset.}
The AP10K-C dataset is derived from the AP10K~\cite{ap10k} dataset, which consists of 10,015 images representing 23 animal families and 54 species, along with their high-quality keypoint annotations. Following the COCO~\cite{coco} style, the AP10K dataset supports a 17-keypoint annotation for animal bodies.

\subsection{Evaluation Metrics}\label{sec:metrics}
In this section, we present two standard evaluation metrics for pose estimation: mean average precision (mAP) and mean average recall (mAR). These metrics evaluate model accuracy on both clean and corrupted test images. Additionally, following ~\cite{robotad}, we introduce the concept of relative robustness to assess a model's robustness.

\paragraph{mAP and mAR.} are two widely adopted metrics for assessing pose estimation performance. Following ~\cite{simplebaseline, hourglass}, we report mAP as the average of AP values at thresholds ranging from 0.5 to 0.95, with a step size of 0.05. We also provide ``AP .5" and ``AP .75", representing the AP at thresholds of 0.5 and 0.75, respectively. ``AP (M)" and ``AP (L)" measure the AP for medium-sized and large-sized objects, respectively. Similarly, we report mAR, AR .5, AR .75, AR (M), and AR (L).

\paragraph{Mean Relative Robustness (mRR).} 
Following~\cite{robotad}, we introduce the robustness metric, mean Relative Robustness (mRR), to evaluate how much a model's performance drops under certain corruptions compared to clean images.
To calculate this metric, we first evaluate the model on clean images and obtain the mean Average Precision (mAP), denoted as $mAP_{clean}$. For any corruption $c$, we then calculate the mAP at each severity level $s$, denoted as $mAP_{c,s}$. The relative robustness for corruption $c$ and the mean Relative Robustness (mRR) are defined as follows:
\begin{equation}
    RR_c = \frac{1}{5} \sum_{s=1}^{5}(1 - \frac{mAP_{clean} - mAP_{c,s}}{mAP_{clean}}), \ \ mRR = \frac{1}{C} \sum_{c=1}^{C} RR_c,
\end{equation}
where $c$ indexes the $C$ types of corruption.

\section{Experiment}\label{sec:experiment}

\begin{figure}
    \centering
    \includegraphics[width=\linewidth]{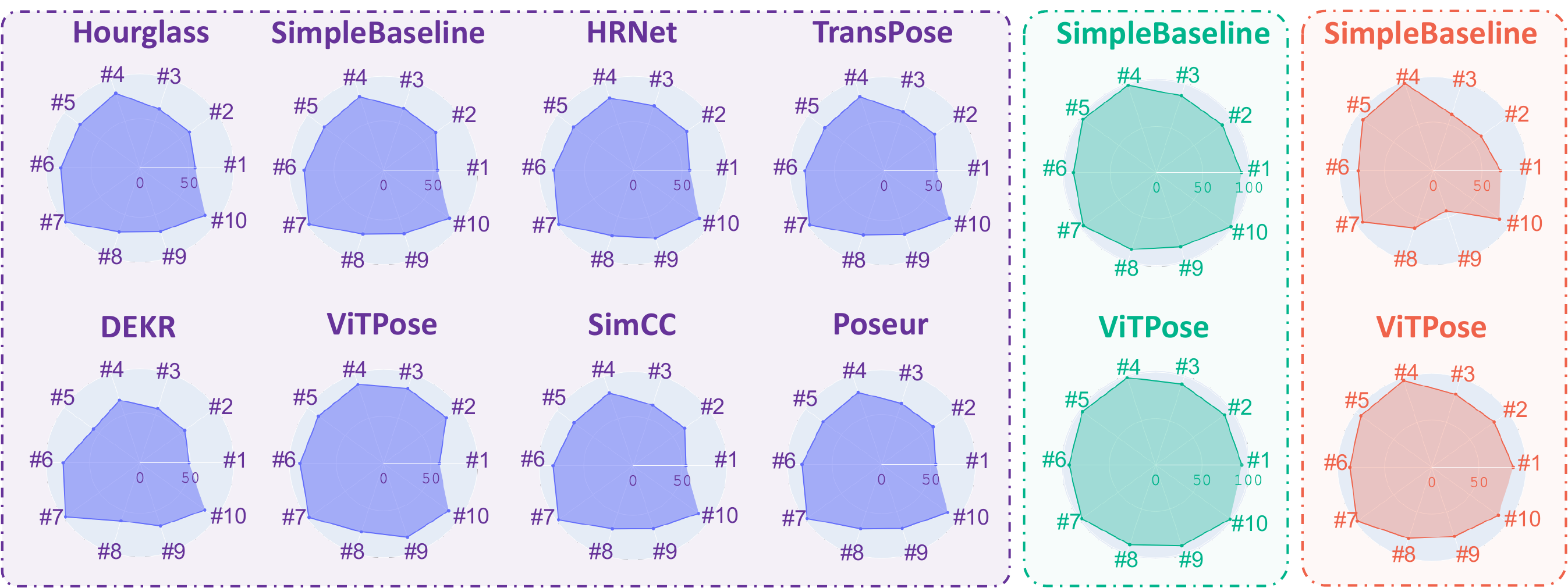}
    \caption{Pose estimation results in terms of mRR (\%) for 10 representative models on the \textcolor{robo_blue}{COCO-C}, \textcolor{robo_green}{OCHuman-C}, and \textcolor{robo_red}{AP10K-C} datasets. \textbf{Corruptions:} \#1 \texttt{Motion Blur}, \#2 \texttt{Gaussian Noise}, \#3 \texttt{Impulse Noise}, \#4 \texttt{Pixelate}, \#5 \texttt{JPEG Compression}, \#6 \texttt{Color Quant}, \#7 \texttt{Brightness}, \#8 \texttt{Darkness}, \#9 \texttt{Contrast}, \#10 \texttt{Mask}.
    }
    \label{fig:model_on_corruptions}
\end{figure}

\begin{figure}
    \centering
    \begin{subfigure}[b]{0.30\textwidth}
         \centering
         \includegraphics[width=\textwidth]{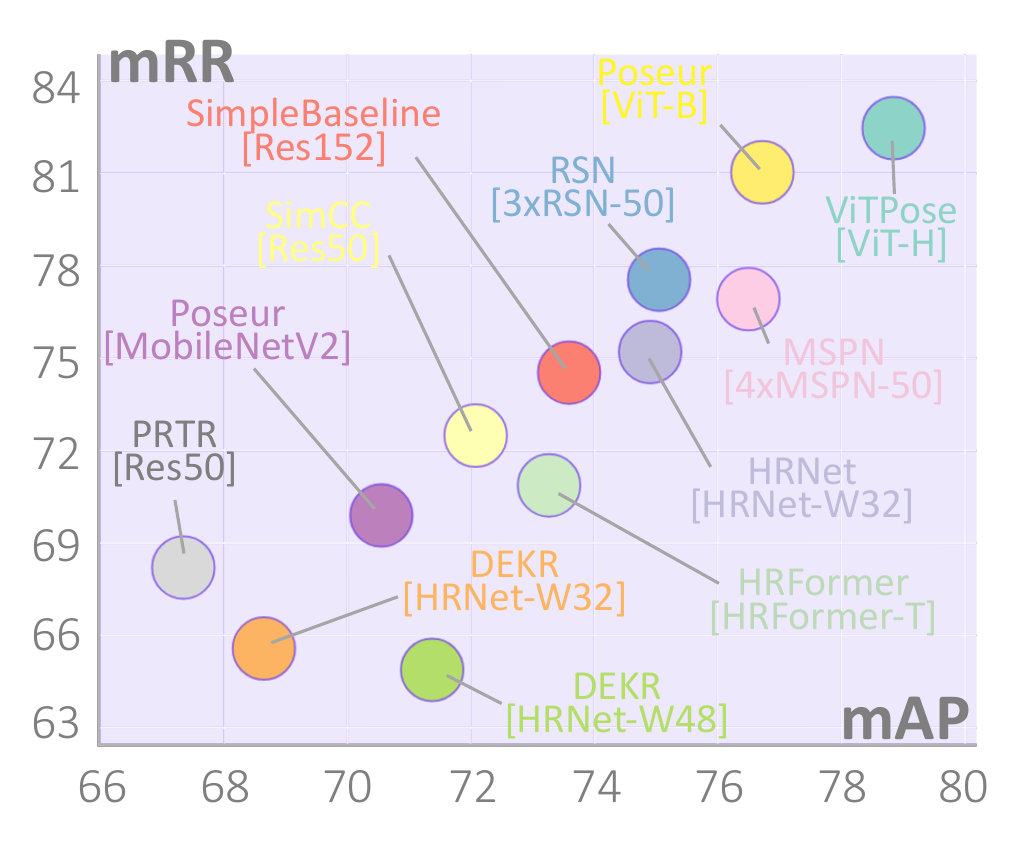}
         \caption{COCO-C}
         \label{fig:coco_clean_ap_vs_rr}
    \end{subfigure}
    \hfill
    \begin{subfigure}[b]{0.307\textwidth}
         \centering
         \includegraphics[width=\textwidth]{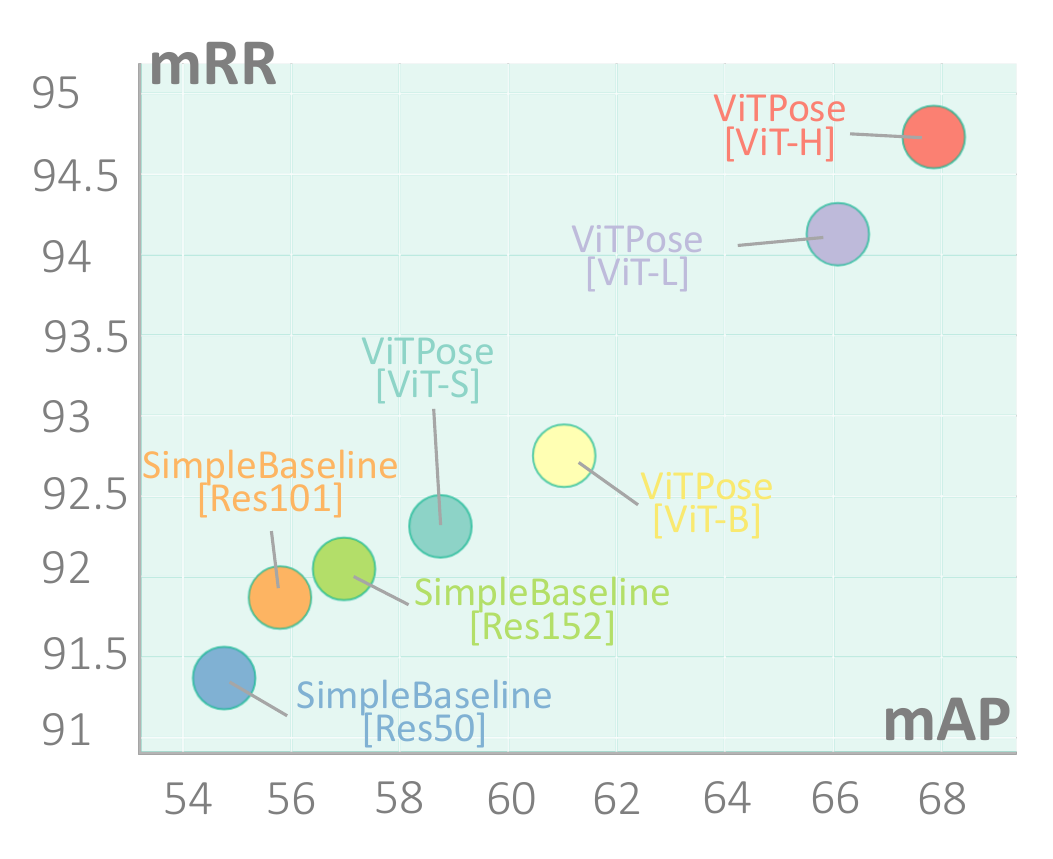}
         \caption{OCHuman-C}
         \label{fig:ochuman_clean_ap_vs_rr}
    \end{subfigure}
    \hfill
    \begin{subfigure}[b]{0.30\textwidth}
         \centering
         \includegraphics[width=\textwidth]{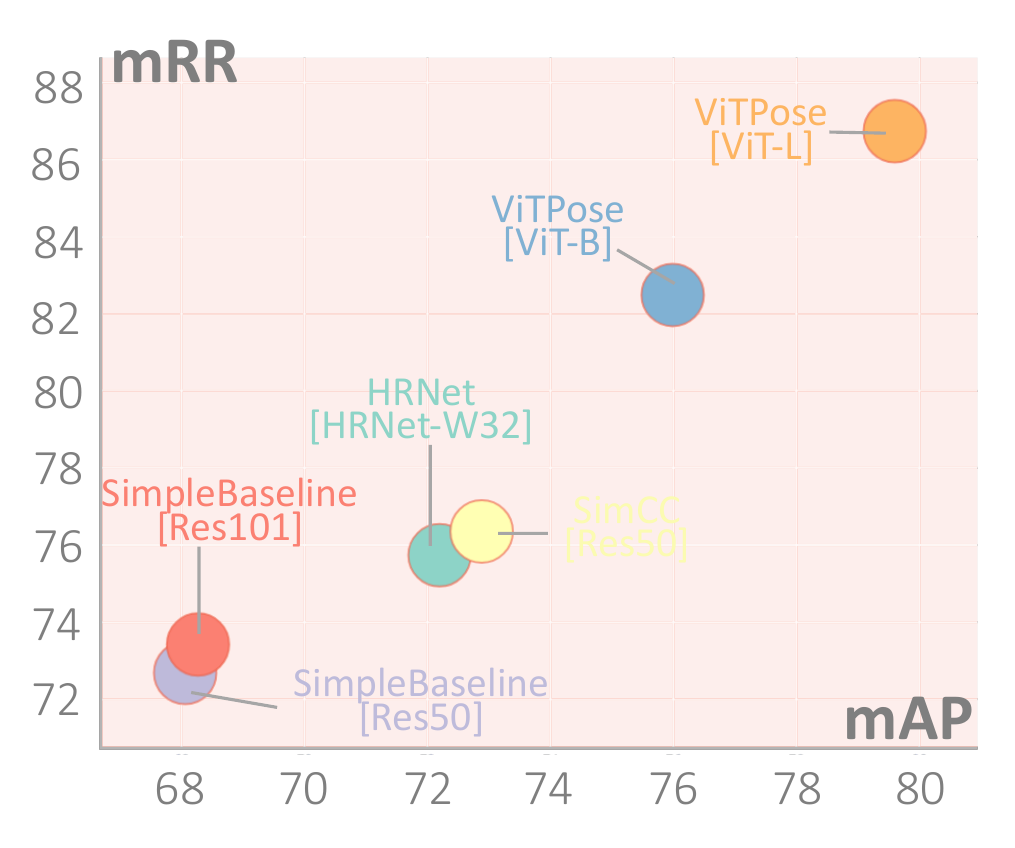}
         \caption{AP10K-C}
         \label{fig:ap10k_clean_ap_vs_corr_ap}
    \end{subfigure}
    \caption{The strong correlation between clean mAP and corruption robustness mRR (\%).}
    \label{fig:robustness_vs_clean_ap}
\end{figure}

\subsection{Benchmark Settings}

\paragraph{Pose Estimation Models.}
In this study, we conduct a comprehensive benchmark of 15 pose estimation methods and their 60 variants, representing the most significant models in the pose estimation domain. Our evaluation encompasses a variety of approaches, including top-down~\cite{hourglass, hrnet, hrformer, vitpose}, bottom-up~\cite{ae, higherhrnet, dekr}, heatmap-based~\cite{transpose, rsn, mspn}, regression-based~\cite{prtr, poseur}, and classification-based~\cite{simcc} methods. The specific models assessed are Hourglass~\cite{hourglass}, SimpleBaseline~\cite{simplebaseline}, HRNet~\cite{hrnet}, MSPN~\cite{mspn}, RSN~\cite{rsn}, TransPose~\cite{transpose}, HRFormer~\cite{hrformer}, LiteHRNet~\cite{litehrnet}, ViTPose~\cite{vitpose}, SimCC~\cite{simcc}, Associative Embedding (AE)~\cite{ae}, HigherHRNet~\cite{higherhrnet}, DEKR~\cite{dekr}, Pose Transformer (PRTR)~\cite{prtr} and Poseur~\cite{poseur}, covering both CNN-based models and ViT-based models.

\paragraph{Datasets.}
We conduct a comprehensive evaluation using three representative datasets: the human body datasets COCO-C and OCHuman-C, and the animal pose estimation dataset AP10K-C (Section~\ref{sec:datasets}). Without further description, all models are trained on the original clean training sets and evaluated on the corrupted validation sets. Notably, for the OCHuman-C dataset, models trained on COCO~\cite{coco} training set are directly evaluated without additional fine-tuning.

\paragraph{Evaluation Protocols.}
We utilize public checkpoints and settings for our evaluations. For most methods, we employ the MMPose open-source framework~\cite{mmpose} and its provided public checkpoints. However, for Poseur~\cite{poseur}, PRTR~\cite{prtr}, HigherHRNet~\cite{higherhrnet}, TransPose~\cite{transpose}, and ViTPose~\cite{vitpose}, the MMPose framework does not offer a complete setup and checkpoints. Therefore, we use the official repositories open-sourced by their authors. We standardize the evaluation by exploring common configurations across different methods, such as backbone architecture, data augmentation, input resolution, pre-training, and post-processing. To ensure a comprehensive evaluation, we report metrics for each corruption type by averaging results over five severity levels.

\subsection{Robustness Analysis}
Table~\ref{tab:coco} presents the results of 15 methods and their 60 variants evaluated on the COCO-C dataset under four different corruption scenarios. The findings reveal that all models, regardless of being top-down or bottom-up, heatmap-based, regression-based, or classification-based, experience varying degrees of performance degradation across different corruption types. Figure~\ref{fig:model_on_corruptions} shows the corruption robustness of 8 representative models on the COCO dataset in terms of mRR (\%).

\begin{table*}[]
    \caption{
    Robustness benchmark results on the COCO-C dataset, with mRR scores presented as percentages (\%). The blocks, from top to bottom, represent top-down, bottom-up, and regression-based methods. All methods in the top-down block are heatmap-based, except for SimCC~\cite{simcc}. \textbf{Bold}: Best in column. \underline{Underline}: Second best in column. The best score of each metric across four corruption sets is highlighted in background color.
    }
    \centering\scalebox{0.51}{
    \begin{tabular}{rc|cc|ccc|ccc|ccc|ccc|ccc}
    \toprule
    \multirow{2}{*}{\textbf{Method}} & \multirow{2}{*}{\textbf{Backbone}} & \multicolumn{2}{c}{\textbf{Clean}}\vline & \multicolumn{3}{c}{\textbf{Overall Robustness}}\vline & \multicolumn{3}{c}{\textbf{Blur \& Noise}}\vline & \multicolumn{3}{c}{\textbf{Compression \& Color}}\vline & \multicolumn{3}{c}{\textbf{Lightning}}\vline & \multicolumn{3}{c}{\textbf{Mask}}
    \\
    ~ & ~ & \textcolor{robo_blue}{\textbf{mAP}} & \textcolor{robo_red}{\textbf{mAR}} & \textcolor{robo_blue}{\textbf{mAP}} & \textcolor{robo_red}{\textbf{mAR}} & \textcolor{robo_green}{\textbf{mRR}} & \textcolor{robo_blue}{\textbf{mAP}} & \textcolor{robo_red}{\textbf{mAR}} & \textcolor{robo_green}{\textbf{mRR}} & \textcolor{robo_blue}{\textbf{mAP}} & \textcolor{robo_red}{\textbf{mAR}} & \textcolor{robo_green}{\textbf{mRR}} & \textcolor{robo_blue}{\textbf{mAP}} & \textcolor{robo_red}{\textbf{mAR}} & \textcolor{robo_green}{\textbf{mRR}} & \textcolor{robo_blue}{\textbf{mAP}} & \textcolor{robo_red}{\textbf{mAR}} & \textcolor{robo_green}{\textbf{mRR}} \\
    \midrule\midrule
    Hourglass~\cite{hourglass} & Hourglass-52 & 72.58 & 77.98 & 52.86 & 58.49 & 72.84 & 43.86 & 49.55 & 60.43 & 57.01 & 62.45 & 78.55 & 55.57 & 61.28 & 76.57 & \cellcolor{blue!10} 59.34 & \cellcolor{red!10} 65.06 & \cellcolor{green!10} 81.76\\
    SimpleBaseline~\cite{simplebaseline} & Res50 & 71.82 & 77.38 & 52.32 & 58.18 & 72.84 & 44.45 & 50.57 & 61.90 & 55.71 & 61.28 & 77.57 & 54.54 & 60.47 & 75.94 & \cellcolor{blue!10} 59.05 & \cellcolor{red!10} 64.84 & \cellcolor{green!10} 82.22\\ 
    SimpleBaseline~\cite{simplebaseline} & Res101 & 72.78 & 78.28 & 53.68 & 59.51 & 73.75 & 45.79 & 51.86 & 62.91 & 57.44 & 62.93 & 78.92 & 55.67 & 61.59 & 76.48 & \cellcolor{blue!10} 60.10 & \cellcolor{red!10} 65.93 & \cellcolor{green!10} 82.57\\ 
    SimpleBaseline~\cite{simplebaseline} & Res152 & 73.59 & 79.09 & 54.84 & 60.63 & 74.53 & 47.23 & 53.21 & 64.18 & 58.48 & 63.95 & 79.47 & 56.76 & 62.62 & 77.13 & \cellcolor{blue!10} 61.03 & \cellcolor{red!10} 66.95 & \cellcolor{green!10} 82.93\\ 
    HRNet~\cite{hrnet} & HRNet-W32 & 74.90 & 80.38 & 56.33 & 61.93 & 75.20 & 48.75 & 54.58 & 65.09 & 58.71 & 64.07 & 78.38 & 59.42 & 65.00 & 79.33 & \cellcolor{blue!10} 62.61 & \cellcolor{red!10} 68.29 & \cellcolor{green!10} 83.60\\ 
    HRNet~\cite{hrnet} & HRNet-W48 & 75.58 & 80.85 & 57.52 & 63.04 & 76.10 & 50.29 & 56.09 & 66.53 & 60.03 & 65.32 & 79.42 & 60.12 & 65.60 & 79.54 & \cellcolor{blue!10} 63.85 & \cellcolor{red!10} 69.42 & \cellcolor{green!10} 84.48\\ 
    MSPN~\cite{mspn} & MSPN-50 & 72.28 & 78.80 & 52.37 & 59.20 & 72.46 & 43.57 & 50.48 & 60.28 & 56.13 & 62.83 & 77.66 & 55.15 & 62.06 & 76.30 & \cellcolor{blue!10} 59.17 & \cellcolor{red!10} 65.89 & \cellcolor{green!10} 81.87\\ 
    MSPN~\cite{mspn} & 2xMSPN-50 & 75.35 & 81.56 & 56.71 & 63.32 & 75.26 & 48.59 & 55.30 & 64.49 & 60.19 & 66.63 & 79.88 & 59.21 & 65.89 & 78.58 & \cellcolor{blue!10} 63.11 & \cellcolor{red!10} 69.72 & \cellcolor{green!10} 83.76\\ 
    MSPN~\cite{mspn} & 3xMSPN-50 & 75.85 & 82.06 & 58.12 & 64.81 & 76.63 & 50.52 & 57.34 & 66.61 & 61.85 & 68.34 & 81.55 & 60.34 & 67.06 & 79.55 & \cellcolor{blue!10} 63.12 & \cellcolor{red!10} 69.85 & \cellcolor{green!10} 83.22\\ 
    MSPN~\cite{mspn} & 4xMSPN-50 & 76.49 & 82.62 & 58.83 & 65.59 & 76.92 & 51.42 & 58.39 & 67.22 & 62.72 & 69.20 & 82.0 & 60.63 & 67.42 & 79.26 & \cellcolor{blue!10} 64.03 & \cellcolor{red!10} 70.83 & \cellcolor{green!10} 83.70\\ 
    RSN~\cite{rsn} & RSN-18 & 70.40 & 77.28 & 51.70 & 58.74 & 73.44 & 43.65 & 50.75 & 62.01 & 55.52 & 62.41 & 78.86 & 54.18 & 61.31 & 76.96 & \cellcolor{blue!10} 56.94 & \cellcolor{red!10} 64.02 & \cellcolor{green!10} 80.89\\ 
    RSN~\cite{rsn} & RSN-50 & 72.41 & 79.07 & 54.36 & 61.41 & 75.08 & 47.03 & 54.30 & 64.95 & 57.89 & 64.68 & 79.95 & 56.59 & 63.70 & 78.15 & \cellcolor{blue!10} 59.09 & \cellcolor{red!10} 66.09 & \cellcolor{green!10} 81.61\\ 
    RSN~\cite{rsn} & 2xRSN-50 & 74.76 & 81.02 & 57.70 & 64.45 & 77.18 & 50.34 & 57.33 & 67.34 & 61.75 & 68.22 & 82.60 & 59.60 & 66.38 & 79.72 & \cellcolor{blue!10} 61.91 & \cellcolor{red!10} 68.69 & \cellcolor{green!10} 82.82\\ 
    RSN~\cite{rsn} & 3xRSN-50 & 75.05 & 81.37 & 58.19 & 64.91 & 77.54 & 50.94 & 57.84 & 67.88 & \cellcolor{blue!10} 62.73 & \cellcolor{red!10} 69.21 & \cellcolor{green!10} 83.59 & 59.50 & 66.25 & 79.28 & 62.42 & 69.19 & 83.18\\ 
    TransPose~\cite{transpose} & Res50-A3 & 71.54 & 77.03 & 51.54 & 57.19 & 72.04 & 42.57 & 48.33 & 59.50 & 55.46 & 60.86 & 77.52 & 54.38 & 60.15 & 76.01 & \cellcolor{blue!10} 58.17 & \cellcolor{red!10} 63.89 & \cellcolor{green!10} 81.30\\ 
    TransPose~\cite{transpose} & Res50-A4 & 72.64 & 77.96 & 52.67 & 58.26 & 72.52 & 43.50 & 49.25 & 59.88 & 56.76 & 62.09 & 78.14 & 55.57 & 61.24 & 76.51 & \cellcolor{blue!10} 59.25 & \cellcolor{red!10} 64.85 & \cellcolor{green!10} 81.57\\ 
    TransPose~\cite{transpose} & HRNet-W32 & 74.17 & 79.45 & 54.93 & 60.44 & 74.06 & 46.43 & 52.11 & 62.60 & 57.97 & 63.26 & 78.16 & 58.29 & 63.82 & 78.59 & \cellcolor{blue!10} 61.24 & \cellcolor{red!10} 66.79 & \cellcolor{green!10} 82.57\\ 
    TransPose~\cite{transpose} & HRNet-W48-A4 & 75.28 & 80.33 & 56.54 & 61.96 & 75.11 & 47.94 & 53.61 & 63.68 & 59.30 & 64.51 & 78.77 & 60.23 & 65.58 & 80.01 & \cellcolor{blue!10} 63.00 & \cellcolor{red!10} 68.50 & \cellcolor{green!10} 83.68\\ 
    TransPose~\cite{transpose} & HRNet-W48-A6 & 75.78 & 80.79 & 56.90 & 62.47 & 75.09 & 48.43 & 54.36 & 63.90 & 59.77 & 65.08 & 78.87 & 60.26 & 65.75 & 79.51 & \cellcolor{blue!10} 63.67 & \cellcolor{red!10} 69.14 & \cellcolor{green!10} 84.01\\ 
    HRFormer~\cite{hrformer} & HRFormer-S & 73.84 & 79.27 & 54.87 & 60.67 & 74.31 & 47.30 & 53.38 & 64.06 & 57.13 & 62.68 & 77.38 & 57.95 & 63.77 & 78.48 & \cellcolor{blue!10} 61.54 & \cellcolor{red!10} 67.21 & \cellcolor{green!10} 83.35\\ 
    HRFormer~\cite{hrformer} & HRFormer-B & 75.37 & 80.67 & 56.67 & 62.31 & 75.19 & 49.34 & 55.28 & 65.46 & 59.04 & 64.48 & 78.34 & 59.34 & 64.88 & 78.73 & \cellcolor{blue!10} 63.55 & \cellcolor{red!10} 69.22 & \cellcolor{green!10} 84.31\\ 
    LiteHRNet~\cite{litehrnet} & LiteHRNet-18 & 64.16 & 70.45 & 45.80 & 52.11 & 71.39 & 37.15 & 43.58 & 57.91 & \cellcolor{blue!10} 51.20 & \cellcolor{red!10} 57.28 & \cellcolor{green!10} 79.81 & 47.57 & 54.01 & 74.15 & 50.20 & 56.48 & 78.25\\ 
    LiteHRNet~\cite{litehrnet} & LiteHRNet-30 & 67.54 & 73.61 & 49.08 & 55.33 & 72.66 & 40.30 & 46.75 & 59.66 & \cellcolor{blue!10} 53.90 & \cellcolor{red!10} 59.85 & \cellcolor{green!10} 79.80 & 51.57 & 57.92 & 76.35 & 53.45 & 59.75 & 79.14\\ 
    ViTPose~\cite{vitpose} & ViT-S & 73.92 & 79.24 & 56.73 & 62.56 & 76.75 & 49.58 & 55.72 & 67.07 & 60.38 & 65.92 & 81.69 & 59.01 & 64.78 & 79.83 & \cellcolor{blue!10} 60.43 & \cellcolor{red!10} 66.29 & \cellcolor{green!10} 81.75\\ 
    ViTPose~\cite{vitpose} & ViT-B & 75.75 & 80.99 & 59.50 & 65.13 & 78.55 & 52.45 & 58.38 & 69.25 & \cellcolor{blue!10} 63.00 & 68.32 & \cellcolor{green!10} 83.17 & 61.91 & 67.53 & 81.73 & 62.90 & \cellcolor{red!10} 68.62 & 83.04\\ 
    ViTPose~\cite{vitpose} & ViT-L & \underline{78.18} & \underline{83.43} & \underline{63.83} & \underline{69.37} & \underline{81.65} & \underline{57.44} & \underline{63.23} & \underline{73.47} & \cellcolor{blue!10} \underline{67.18} & \cellcolor{red!10} \underline{72.42} & \cellcolor{green!10} \underline{85.94} & \underline{66.01} & \underline{71.56} & {84.43} & 66.43 & 72.08 & 84.97\\ 
    ViTPose~\cite{vitpose} & ViT-H & \textbf{78.84} & \textbf{83.92} & \textbf{65.02} & \textbf{70.39} & \textbf{82.46} & \textbf{58.89} & \textbf{64.56} & \textbf{74.7} & \cellcolor{blue!10} \textbf{68.56} & \cellcolor{red!10} \textbf{73.65} & \cellcolor{green!10} \textbf{86.96} & \textbf{66.96} & \textbf{72.29} & \textbf{84.93} & 66.93 & 72.37 & 84.89\\ 
    SimCC~\cite{simcc} & Res50 & 72.07 & 78.16 & 52.25 & 58.36 & 72.49 & 43.98 & 50.20 & 61.01 & 55.84 & 61.69 & 77.47 & 54.61 & 60.78 & 75.77 & \cellcolor{blue!10} 59.17 & \cellcolor{red!10} 65.53 & \cellcolor{green!10} 82.10\\ 
    SimCC~\cite{simcc} & MobileNetV2 & 61.98 & 67.81 & 41.45 & 47.05 & 66.88 & 32.74 & 38.33 & 52.82 & 45.72 & 51.19 & 73.78 & 43.41 & 49.08 & 70.04 & \cellcolor{blue!10} 48.89 & \cellcolor{red!10} 54.71 & \cellcolor{green!10} 78.88\\ 
    SimCC~\cite{simcc} & vipnas-mbv3 & 69.48 & 75.52 & 50.31 & 56.50 & 72.41 & 42.06 & 48.45 & 60.54 & 53.90 & 59.81 & 77.57 & 53.0 & 59.23 & 76.28 & \cellcolor{blue!10} 56.23 & \cellcolor{red!10} 62.54 & \cellcolor{green!10} 80.93\\ 
    \midrule
    AE~\cite{ae} & HRNet-W32 & 65.55 & 71.12 & 42.53 & 47.56 & 64.89 & 35.46 & 40.81 & 54.11 & 41.49 & 46.01 & 63.30 & 47.21 & 52.20 & 72.02 & \cellcolor{blue!10} 52.82 & \cellcolor{red!10} 58.53 & \cellcolor{green!10} 80.59\\ 
    HigherHRNet~\cite{higherhrnet} & HRNet-W32 & 67.66 & 72.38 & 43.71 & 48.15 & 64.60 & 35.86 & 40.50 & 53.00 & 43.21 & 47.24 & 63.87 & 48.63 & 53.06 & 71.88 & \cellcolor{blue!10} 53.95 & \cellcolor{red!10} 59.14 & \cellcolor{green!10} 79.74\\ 
    HigherHRNet~\cite{higherhrnet} & HRNet-W48 & 68.51 & 73.07 & 44.91 & 49.17 & 65.55 & 38.04 & 42.82 & 55.52 & 43.79 & 47.44 & 63.91 & 49.53 & 53.68 & 72.30 & \cellcolor{blue!10} 55.00 & \cellcolor{red!10} 59.86 & \cellcolor{green!10} 80.28\\ 
    DEKR~\cite{dekr} & HRNet-W32 & 68.64 & 73.49 & 45.01 & 50.11 & 65.57 & 36.67 & 42.17 & 53.42 & 45.40 & 50.11 & 66.13 & 49.78 & 54.70 & 72.52 & \cellcolor{blue!10} 54.57 & \cellcolor{red!10} 60.12 & \cellcolor{green!10} 79.49\\ 
    DEKR~\cite{dekr} & HRNet-W48 & 71.37 & 76.23 & 46.30 & 51.46 & 64.88 & 38.91 & 44.61 & 54.52 & 44.79 & 49.37 & 62.76 & 51.72 & 56.70 & 72.47 & \cellcolor{blue!10} 56.77 & \cellcolor{red!10} 62.57 & \cellcolor{green!10} 79.54\\ 
    \midrule
    PRTR~\cite{prtr} & Res50 & 67.34 & 74.87 & 45.93 & 54.21 & 68.20 & 38.05 & 46.56 & 56.50 & 47.81 & 56.04 & 71.01 & 49.14 & 57.27 & 72.97 & \cellcolor{blue!10} 54.29 & \cellcolor{red!10} 62.48 & \cellcolor{green!10} 80.62\\ 
    PRTR~\cite{prtr} & Res101 & 68.95 & 76.36 & 48.02 & 56.36 & 69.65 & 40.58 & 49.20 & 58.85 & 49.73 & 58.11 & 72.13 & 51.11 & 59.18 & 74.13 & \cellcolor{blue!10} 55.98 & \cellcolor{red!10} 64.15 & \cellcolor{green!10} 81.20\\ 
    PRTR~\cite{prtr} & HRNet-W32 & 71.71 & 78.30 & 51.21 & 58.65 & 71.42 & 43.70 & 51.36 & 60.94 & 53.06 & 60.53 & 74.00 & 54.37 & 61.56 & 75.82 & \cellcolor{blue!10} 58.75 & \cellcolor{red!10} 66.12 & \cellcolor{green!10} 81.93\\ 
    Poseur~\cite{poseur} & MobileNetV2 & 70.55 & 76.19 & 49.31 & 55.12 & 69.89 & 39.64 & 45.46 & 56.19 & 53.97 & 59.64 & 76.51 & 51.89 & 57.80 & 73.55 & \cellcolor{blue!10} 56.56 & \cellcolor{red!10} 62.55 & \cellcolor{green!10} 80.17\\ 
    Poseur~\cite{poseur} & Res50 & 74.22 & 79.59 & 53.79 & 59.57 & 72.47 & 45.49 & 51.45 & 61.29 & 56.74 & 62.29 & 76.45 & 56.60 & 62.41 & 76.26 & \cellcolor{blue!10} 61.40 & \cellcolor{red!10} 67.28 & \cellcolor{green!10} 82.72\\ 
    Poseur~\cite{poseur} & HRNet-W32 & 75.52 & 80.65 & 55.91 & 61.53 & 74.04 & 47.45 & 53.33 & 62.84 & 58.17 & 63.59 & 77.03 & 59.56 & 65.09 & 78.87 & \cellcolor{blue!10} 63.57 & \cellcolor{red!10} 69.26 & \cellcolor{green!10} 84.17\\ 
    Poseur~\cite{poseur} & HRNet-W48 & 77.62 & 82.33 & 59.97 & 65.03 & 77.26 & 52.31 & 57.60 & 67.39 & 60.61 & 65.57 & 78.09 & 64.22 & 69.11 & 82.75 & \cellcolor{blue!10} \underline{68.26} & \cellcolor{red!10} 73.41 & \cellcolor{green!10} \underline{87.95}\\ 
    Poseur~\cite{poseur} & HRFormer-T & 73.26 & 78.71 & 51.92 & 57.72 & 70.87 & 42.56 & 48.50 & 58.09 & 54.56 & 60.17 & 74.47 & 55.97 & 61.76 & 76.40 & \cellcolor{blue!10} 59.95 & \cellcolor{red!10} 65.90 & \cellcolor{green!10} 81.83\\ 
    Poseur~\cite{poseur} & HRFormer-S & 75.49 & 80.78 & 55.95 & 61.79 & 74.12 & 47.69 & 53.87 & 63.18 & 58.21 & 63.83 & 77.11 & 59.47 & 65.19 & 78.78 & \cellcolor{blue!10} 63.40 & \cellcolor{red!10} 69.27 & \cellcolor{green!10} 83.99\\ 
    Poseur~\cite{poseur} & HRFormer-B & {77.97} & {82.95} & 61.84 & 67.23 & 79.32 & 54.48 & 60.18 & 69.88 & 63.27 & 68.52 & 81.15 & {65.54} & {70.73} & 84.07 & \cellcolor{blue!10} \textbf{68.53} & \cellcolor{red!10} \textbf{74.05} & \cellcolor{green!10} 87.90\\ 
    Poseur~\cite{poseur} & ViT-B & 76.72 & 81.92 & {62.17} & {67.54} & {81.03} & {54.61} & {60.24} & {71.19} & {64.83} & {69.95} & {84.51} & 65.11 & 70.44 & \underline{84.87} & \cellcolor{blue!10} 67.99 & \cellcolor{red!10} \underline{73.52} & \cellcolor{green!10} \textbf{88.63}\\ 
    \bottomrule
    \end{tabular}
    }
\label{tab:coco}
\end{table*}

\begin{figure}[t]
    \centering
    \includegraphics[width=\linewidth]{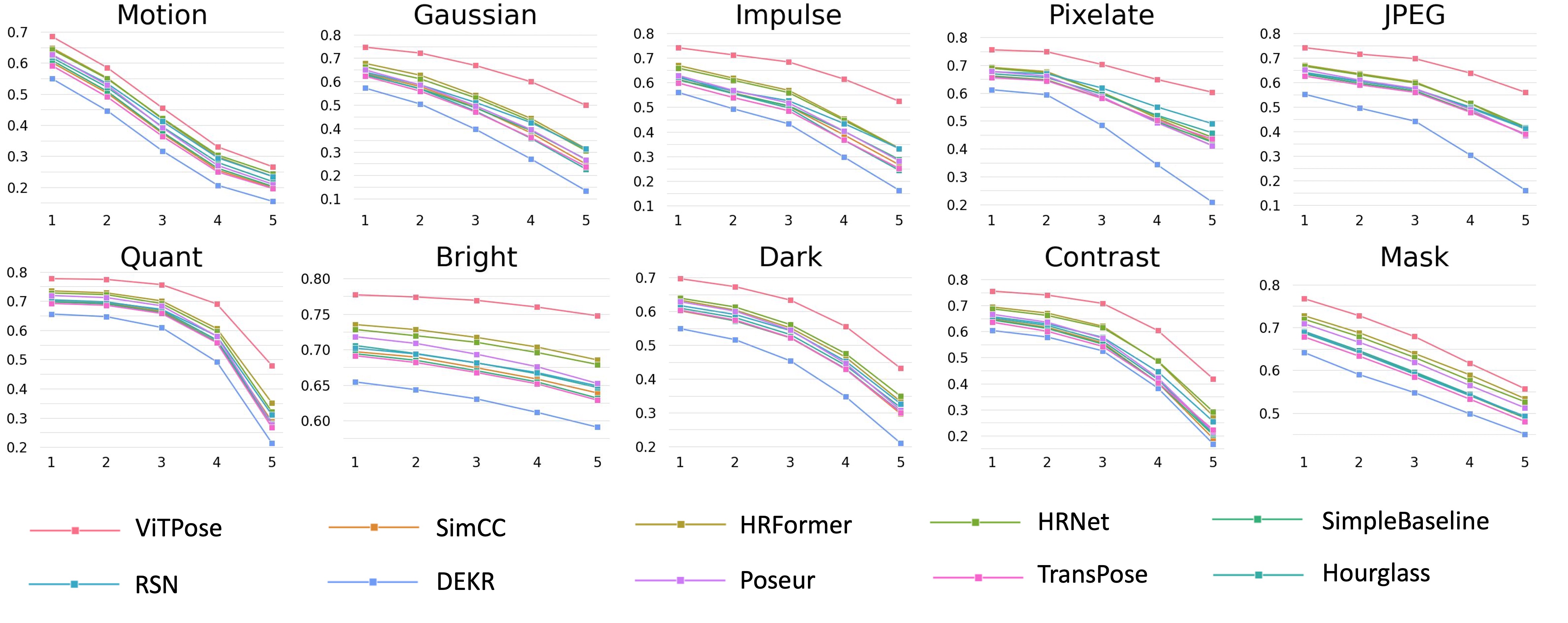}
    \caption{
    Per-Severity Error Analysis: The mRR (\%) results for 10 representative methods on the COCO-C dataset, with higher severity levels indicating greater degrees of corruption.
    }
    \label{fig:per_sev_err}
\end{figure}

The overall relative robustness of models against all corruptions is strongly correlated with their performance on original clean images, as shown in Figure~\ref{fig:coco_clean_ap_vs_rr}. In Table~\ref{tab:coco}, models that perform better on clean images, such as ViTPose-L and ViTPose-H~\cite{vitpose}, tend to exhibit the best robustness under corrupted conditions. A deeper analysis of per-severity error rates, illustrated in Figure~\ref{fig:per_sev_err}, supports this conclusion. For instance, ViTPose-H~\cite{vitpose}, with a mAP of 78.84, demonstrates higher corruption robustness than DEKR HRNet-W32~\cite{dekr}, which has a mAP of 68.64, across all severity levels. As the severity level increases, all models show a greater degradation in corruption robustness, as indicated by the mRR(\%) values.

\begin{figure}
    \centering
    \begin{subfigure}[b]{0.30\textwidth}
         \centering
         \includegraphics[width=\textwidth]{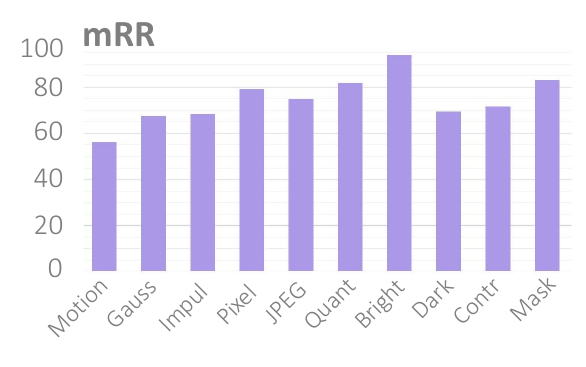}
         \caption{COCO-C}
         \label{fig:coco_rr_corruptions}
    \end{subfigure}
    \hfill
    \begin{subfigure}[b]{0.30\textwidth}
         \centering
         \includegraphics[width=\textwidth]{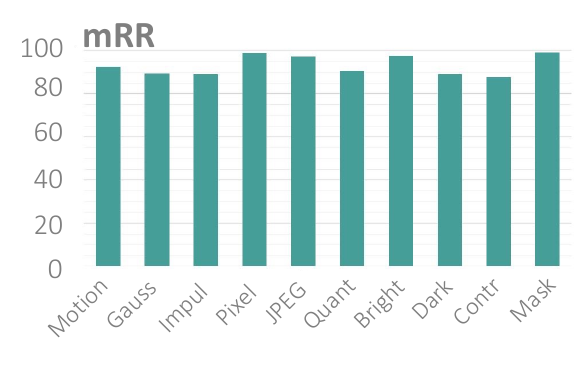}
         \caption{OCHuman-C}
         \label{fig:ochuman_rr_corruptions}
    \end{subfigure}
    \hfill
    \begin{subfigure}[b]{0.30\textwidth}
         \centering
         \includegraphics[width=\textwidth]{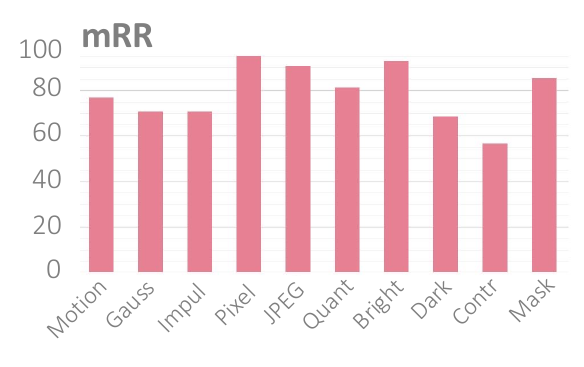}
         \caption{AP10K-C}
         \label{fig:ap10k_rr_corruptions}
    \end{subfigure}
    \caption{
    Averaged mRR values of all models across various corruptions for each dataset.}
    \label{fig:robustness_on_corruptions}
\end{figure}

\begin{table}[!t]
    \caption{
    Evaluation results of heatmap-based and regression-based methods on the COCO-C dataset under Mask corruption. Metrics are calculated on corrupted images, except for ``Clean mAP'', which is measured on the original images. The best and second-best results in each column are highlighted in \textbf{bold} and \underline{underline}, respectively.
    }
    \centering\scalebox{0.64}{
    \begin{tabular}{ccc|ccccccccccc}
    \toprule
    \textbf{Model} & \textbf{Backbone} & \textbf{Clean mAP} & \textbf{mAP} & \textbf{AP .5} & \textbf{AP .75} & \textbf{AP (M)} & \textbf{AP (L)} & \textbf{AR} & \textbf{AR .5} & \textbf{AR .75} & \textbf{AR (M)} & \textbf{AR (L)} & \textbf{mRR} \\
    \midrule\midrule
    ViTPose~\cite{vitpose} & ViT-L & \underline{78.18} & 66.43 & 84.68 & 72.30 & 56.93 & \underline{80.96} & 72.08 & 89.13 & 77.23 & 61.79 & 86.43 & 84.97 \\
    ViTPose~\cite{vitpose} & ViT-H & \textbf{78.84} & 66.93 & 84.91 & 72.55 & 57.02 & \textbf{82.01} & 72.37 & 89.17 & 77.28 & 61.72 & \textbf{87.23} & 84.89 \\
    \midrule
    Poseur~\cite{poseur} & HRFormer-B & 77.97 & \textbf{68.53} & \textbf{85.95} & \underline{74.37} & \textbf{60.47} & 80.90 & \textbf{74.05} & \underline{89.88} & \underline{79.10} & \textbf{65.14} & \underline{86.52} & \underline{87.90} \\
    Poseur~\cite{poseur} & ViT-B & 76.72 & \underline{67.99} & \underline{85.69} & \textbf{74.57} & \underline{60.38} & 80.34 & \underline{73.52} & \textbf{90.10} & \textbf{79.22} & \underline{65.08} & 85.43 & \textbf{88.63} \\
    \bottomrule
    \end{tabular}
    }
    \label{tab:robustness_on_mask}
\end{table}

\paragraph{Comparison of Corruption Types.}
Table~\ref{tab:coco} shows the performance decline of various methods across different corruption types. Robustness against each corruption type is closely correlated to the clean AP, except in the mask category. Notably, for regression methods like PRTR~\cite{prtr} and Poseur~\cite{poseur}, despite their lower clean mAP compared to ViTPose~\cite{vitpose}, they outperform ViTPose and show higher robustness in the presence of mask corruption. Table~\ref{tab:robustness_on_mask} provides a detailed comparison of two heatmap-based models (\textit{i.e.}, ViTPose with the ViT-L and ViT-H backbones) and two regression-based models (\textit{i.e.}, Poseur with the HRFormer-B and ViT-B backbones) under mask corruption. Generally, regression-based models excel over heatmap-based ones under mask corruption across most metrics, even though the reverse is true for clean images. This is possible because regression-based methods, by directly predicting coordinates, can better integrate global context information, aiding in the resolution of ambiguities and occlusions.

In Figure~\ref{fig:coco_rr_corruptions}, we show the averaged mRR values of all models on the COCO-C dataset for each corruption type. Motion blur, Gaussian noise, and Impulse Noise cause the most significant performance degradation, while Brightness and Mask corruptions have a milder impact. This difference is likely because images with abundant global context support pose estimation models more effectively. Local corruptions like masks, which maintain most of the contextual information, lead to smaller performance drops compared to global corruptions like motion blur and Gaussian noise. Although brightness is a global corruption, overexposure does not substantially diminish contextual information, resulting in better robustness against this type of corruption.

\begin{table*}[]
    \caption{
    Robustness benchmark results for the OCHuman-C dataset, with mRR values shown as percentages (\%). \textbf{Bold} indicates the best in each column, while the top score for each metric across four corruption sets is highlighted with a unique background color.
    }
    \centering\scalebox{0.53}{
    \begin{tabular}{rc|cc|ccc|ccc|ccc|ccc|ccc}
    \toprule
    \multirow{2}{*}{\textbf{Method}} & \multirow{2}{*}{\textbf{Backbone}} & \multicolumn{2}{c}{\textbf{Clean}}\vline & \multicolumn{3}{c}{\textbf{Overall Robustness}}\vline & \multicolumn{3}{c}{\textbf{Blur \& Noise}}\vline & \multicolumn{3}{c}{\textbf{Compression \& Color}}\vline & \multicolumn{3}{c}{\textbf{Lightning}}\vline & \multicolumn{3}{c}{\textbf{Mask}}
    \\
    ~ & ~ & \textcolor{robo_blue}{\textbf{mAP}} & \textcolor{robo_red}{\textbf{mAR}} & \textcolor{robo_blue}{\textbf{mAP}} & \textcolor{robo_red}{\textbf{mAR}} & \textcolor{robo_green}{\textbf{mRR}} & \textcolor{robo_blue}{\textbf{mAP}} & \textcolor{robo_red}{\textbf{mAR}} & \textcolor{robo_green}{\textbf{mRR}} & \textcolor{robo_blue}{\textbf{mAP}} & \textcolor{robo_red}{\textbf{mAR}} & \textcolor{robo_green}{\textbf{mRR}} & \textcolor{robo_blue}{\textbf{mAP}} & \textcolor{robo_red}{\textbf{mAR}} & \textcolor{robo_green}{\textbf{mRR}} & \textcolor{robo_blue}{\textbf{mAP}} & \textcolor{robo_red}{\textbf{mAR}} & \textcolor{robo_green}{\textbf{mRR}} \\
    \midrule\midrule
    SimpleBaseline~\cite{simplebaseline} & Res50 & 54.76 & 59.30 & 50.04 & 55.03 & 91.37 & 48.10 & 53.35 & 87.84 & 51.74 & 56.52 & 94.48 & 48.80 & 53.91 & 89.11 & \cellcolor{blue!10} 54.45 & \cellcolor{red!10} 59.01 & \cellcolor{green!10} 99.44\\ 
    SimpleBaseline~\cite{simplebaseline} & Res101 & 55.79 & 60.64 & 51.26 & 56.33 & 91.87 & 49.29 & 54.55 & 88.34 & 53.19 & 58.03 & 95.34 & 49.87 & 55.07 & 89.38 & \cellcolor{blue!10} 55.52 & \cellcolor{red!10} 60.33 & \cellcolor{green!10} \textbf{99.52}\\ 
    SimpleBaseline~\cite{simplebaseline} & Res152 & 56.97 & 61.53 & 52.44 & 57.30 & 92.05 & 50.56 & 55.62 & 88.74 & 54.42 & 59.08 & 95.52 & 50.99 & 55.95 & 89.50 & \cellcolor{blue!10} 56.52 & \cellcolor{red!10} 61.03 & \cellcolor{green!10} 99.20\\ 
    ViTPose~\cite{vitpose} & ViT-S & 58.75 & 62.92 & 54.23 & 58.82 & 92.31 & 52.47 & 57.24 & 89.30 & 55.77 & 60.15 & 94.93 & 53.19 & 57.91 & 90.53 & \cellcolor{blue!10} 58.09 & \cellcolor{red!10} 62.28 & \cellcolor{green!10} 98.87\\ 
    ViTPose~\cite{vitpose} & ViT-B & 61.03 & 65.12 & 56.61 & 61.15 & 92.75 & 54.92 & 59.52 & 89.98 & 58.00 & 62.40 & 95.03 & 55.71 & 60.39 & 91.28 & \cellcolor{blue!10} 60.19 & \cellcolor{red!10} 64.58 & \cellcolor{green!10} 98.62\\ 
    ViTPose~\cite{vitpose} & ViT-L & 66.08 & 69.48 & 62.20 & 66.09 & 94.13 & 60.73 & 64.69 & 91.90 & 63.43 & 67.20 & 95.98 & 61.36 & 65.36 & 92.86 & \cellcolor{blue!10} 65.46 & \cellcolor{red!10} 69.20 & \cellcolor{green!10} 99.06\\ 
    ViTPose~\cite{vitpose} & ViT-H & \textbf{67.85} & \textbf{71.04} & \textbf{64.28} & \textbf{67.74} & \textbf{94.73} & \textbf{62.65} & \textbf{66.23} & \textbf{92.33} & \textbf{65.61} & \textbf{68.98} & \textbf{96.70} & \textbf{63.50} & \textbf{67.06} & \textbf{93.58} & \cellcolor{blue!10} \textbf{67.50} & \cellcolor{red!10} \textbf{70.61} & \cellcolor{green!10} 99.47\\ 
    \bottomrule
    \end{tabular}
    }
\label{tab:ochuman}
\end{table*}
\begin{table*}[t!]
    \caption{
    Robustness benchmark results for the AP10K-C dataset, with mRR values shown as percentages (\%). \textbf{Bold} indicates the best in each column, while the top score for each metric across four corruption sets is highlighted with a unique background color.
    }
    \centering\scalebox{0.53}{
    \begin{tabular}{rc|cc|ccc|ccc|ccc|ccc|ccc}
    \toprule
    \multirow{2}{*}{\textbf{Method}} & \multirow{2}{*}{\textbf{Backbone}} & \multicolumn{2}{c}{\textbf{Clean}}\vline & \multicolumn{3}{c}{\textbf{Overall Robustness}}\vline & \multicolumn{3}{c}{\textbf{Blur \& Noise}}\vline & \multicolumn{3}{c}{\textbf{Compression \& Color}}\vline & \multicolumn{3}{c}{\textbf{Lightning}}\vline & \multicolumn{3}{c}{\textbf{Mask}}
    \\
    ~ & ~ & \textcolor{robo_blue}{\textbf{mAP}} & \textcolor{robo_red}{\textbf{mAR}} & \textcolor{robo_blue}{\textbf{mAP}} & \textcolor{robo_red}{\textbf{mAR}} & \textcolor{robo_green}{\textbf{mRR}} & \textcolor{robo_blue}{\textbf{mAP}} & \textcolor{robo_red}{\textbf{mAR}} & \textcolor{robo_green}{\textbf{mRR}} & \textcolor{robo_blue}{\textbf{mAP}} & \textcolor{robo_red}{\textbf{mAR}} & \textcolor{robo_green}{\textbf{mRR}} & \textcolor{robo_blue}{\textbf{mAP}} & \textcolor{robo_red}{\textbf{mAR}} & \textcolor{robo_green}{\textbf{mRR}} & \textcolor{robo_blue}{\textbf{mAP}} & \textcolor{robo_red}{\textbf{mAR}} & \textcolor{robo_green}{\textbf{mRR}} \\
    \midrule\midrule
    SimpleBaseline~\cite{simplebaseline} & Res50 & 68.06 & 71.77 & 49.47 & 53.51 & 72.68 & 43.16 & 47.33 & 63.41 & \cellcolor{blue!10} 58.87 & \cellcolor{red!10} 62.76 & \cellcolor{green!10} 86.50 & 43.84 & 48.0 & 64.41 & 57.08 & 60.91 & 83.87\\ 
    SimpleBaseline~\cite{simplebaseline} & Res101 & 68.27 & 71.92 & 50.12 & 53.99 & 73.42 & 44.85 & 48.90 & 65.69 & \cellcolor{blue!10} 59.38 & \cellcolor{red!10} 63.16 & \cellcolor{green!10} 86.98 & 43.73 & 47.48 & 64.05 & 57.34 & 61.29 & 83.99\\ 
    ViTPose~\cite{vitpose} & ViT-B & 75.99 & 79.41 & 62.68 & 66.34 & 82.49 & 59.78 & 63.59 & 78.67 & \cellcolor{blue!10} 68.63 & \cellcolor{red!10} 72.18 & \cellcolor{green!10} 90.32 & 58.76 & 62.30 & 77.34 & 65.31 & 69.16 & 85.95\\ 
    ViTPose~\cite{vitpose} & ViT-L & \textbf{79.60} & \textbf{82.63} & \textbf{69.04} & \textbf{72.53} & \textbf{86.74} & \textbf{66.21} & \textbf{69.80} & \textbf{83.18} & \cellcolor{blue!10} \textbf{73.37} & \cellcolor{red!10} \textbf{76.71} & \cellcolor{green!10} \textbf{92.18} & \textbf{67.47} & \textbf{70.95} & \textbf{84.76} & \textbf{69.31} & \textbf{72.92} & \textbf{87.07}\\ 
    \bottomrule
    \end{tabular}
    }
\label{tab:ap10k}
\end{table*}

\paragraph{Comparison on Different Datasets.}
Both COCO-C and OCHuman-C are human pose datasets, but the robustness of pose estimation models to corruption on them differs significantly. As shown in Table~\ref{tab:ochuman} and Figure~\ref{fig:ochuman_rr_corruptions}, models exhibit higher overall robustness on OCHuman-C. This is likely because models were tested directly on OCHuman-C without prior training, resulting in a naturally lower clean mAP. While models on both datasets show similar insensitivity to changes in brightness, OCHuman-C models demonstrate notable improvements in robustness against mask and motion blur corruptions. This is probably due to OCHuman-C's high level of occlusion from multi-person interactions, making additional mask and blur corruptions less impactful.

In animal pose estimation, as validated in Table~\ref{tab:ap10k}, models show less sensitivity to image compression corruptions like pixelate and JPEG compression on the AP10K-C dataset. However, they suffer significant performance drops under contrast changes. This likely occurs because the monochromatic skin and fur of animals make these pixel-level losses less impactful. Furthermore, the similarity between animal coloration and natural environments results in poor performance under low-light conditions, such as dark and contrast corruptions.

\subsection{Robustness Enhancement}\label{sec:robust_enhance}
In this section, we examine various enhancement factors to improve the corruption robustness of pose estimation models, aiming to offer valuable insights for training robust pose estimation models.

\paragraph{Impact of Backbones.}
To assess the robustness of various architectural designs, we conducted a thorough evaluation across multiple backbones, as detailed in Table~\ref{tab:coco}. Our findings reveal that models utilizing vision transformer (ViT) backbones, including HRFormer~\cite{hrformer} and ViTPose~\cite{vitpose}, consistently outperform CNN-based counterparts. As shown in Tables~\ref{tab:coco}, \ref{tab:ochuman}, \ref{tab:ap10k}, the best and second best models on clean images and overall corruption robustness are all ViT-based. Specifically, ViTPose~\cite{vitpose} with the ViT-H backbone achieves an impressive 78.84 mAP on clean images and 65.02 mAP on corrupted images, significantly outperforming other models. Notably, when comparing methods available with both CNN and vision transformer backbones, such as Poseur~\cite{poseur}, the transformer backbone (ViT-B) exhibits superior performance over the CNN ones in both mAP and mRR under corruptions, underscoring its significant advantage in robustness. In summary, ViT backbones are more promising for achieving top performance in pose estimation. The advantage stems from the scalability of vision transformer models, which can be easily expanded in terms of layers, hidden units, and attention heads \cite{zhai2022scaling, vaswani2017attention, dosovitskiy2020image}. This scalability enhances the representation capability of large models, enabling them to effectively learn from extensive and intricate datasets, thereby achieving superior performance. In contrast, CNN-based methods often face performance limitations with increasing layer depth. For example, in Table~\ref{tab:coco}, MSPN \cite{mspn} and RSN \cite{rsn} use stacked basic blocks to increase model capacity but show limited robustness, particularly in terms of corruption resilience.

\begin{table}[]
    \caption{Evaluation results of different input resolutions and post processing on COCO-C dataset.}
    \begin{subtable}{0.5\textwidth}
    \caption{Input resolution}
    \centering\scalebox{0.5}{
    \begin{tabular}{rc|c|ccc}
    \toprule
    \textbf{Method}  & \textbf{Backbone} & \textbf{Input size} & \textbf{Clean mAP} & \textbf{Corr mAP} & \textbf{mRR (\%)} \\
    \midrule\midrule
    SimpleBaseline~\cite{simplebaseline} & Res50 & 256x192 & 71.82 & 52.32 & 72.84 \\
    SimpleBaseline~\cite{simplebaseline} & Res50 & 384x288 & 73.06 & 51.16 & 70.03 \\
    \midrule
    SimpleBaseline~\cite{simplebaseline} & Res152 & 256x192 & 73.59 & 54.84 & 74.53 \\
    SimpleBaseline~\cite{simplebaseline} & Res152 & 384x288 & 75.07 & 55.28 & 73.64 \\
    \midrule
    HRNet~\cite{hrnet} & HRNet-W32 & 256x192 & 74.90 & 56.33 & 75.2 \\
    HRNet~\cite{hrnet} & HRNet-W32 & 384x288 & 76.09 & 57.25 & 75.23 \\
    \midrule
    HRNet~\cite{hrnet} & HRNet-W48 & 256x192 & 75.58 & 57.52 & 76.10\\
    HRNet~\cite{hrnet} & HRNet-W48 & 384x288 & 76.71 & 57.68 & 75.20\\
    \bottomrule
    \end{tabular}
    }
    \label{tab:input_size}
    \end{subtable}
    \hspace{\fill}
    \begin{subtable}{0.5\textwidth}
    \caption{Post-processing.}
    \centering\scalebox{0.5}{
    \begin{tabular}{rc|c|ccc}
    \toprule
    \textbf{Method}  & \textbf{Backbone} & \textbf{Post Processing} & \textbf{Clean mAP} & \textbf{Corr mAP} & \textbf{mRR (\%)} \\
    \midrule\midrule
    HRNet~\cite{hrnet} & HRNet-W32 & - & 74.90 & 56.33 & 75.20 \\
    HRNet~\cite{hrnet} & HRNet-W32 & UDP & 76.11 & 57.22 & 75.18 \\
    HRNet~\cite{hrnet} & HRNet-W32 & DARK & 75.72 & 57.00 & 75.28 \\
    \midrule
    HRNet~\cite{hrnet} & HRNet-W48 & - & 75.58 & 57.52 & 76.10 \\
    HRNet~\cite{hrnet} & HRNet-W48 & UDP & 76.76 & 58.23 & 75.87 \\
    HRNet~\cite{hrnet} & HRNet-W48 & DARK & 76.38 & 58.01 & 75.95 \\
    \bottomrule
    \end{tabular}
    }
    \label{tab:post_process}
    \end{subtable}
\end{table}

\begin{table}[]
    \caption{Evaluation results of ViTPose (ViT-B) with different pretraining datasets on COCO-C dataset.}
    \centering\scalebox{0.7}{
    \begin{tabular}{l|lll}
    \toprule
    \textbf{Pretrain Datasets}  & \textbf{Clean mAP} & \textbf{Corr mAP} & \textbf{mRR (\%)} \\
    \midrule\midrule
    COCO~\cite{coco} & 75.75 & 59.50 & 78.55\\
    \midrule
    COCO + AIC~\cite{aic} + MPII~\cite{mpii} & 77.08 (1.33 $\uparrow$) & 60.60 (1.10 $\uparrow$) & 78.62 (0.07 $\uparrow$)\\
    COCO + AIC~\cite{aic} + MPII~\cite{mpii} + CrowdPose~\cite{crowdpose} & 77.51 (1.76 $\uparrow$) & 60.91 (1.41 $\uparrow$) & 78.59 (0.04 $\uparrow$)\\
    COCO + AIC~\cite{aic} + MPII~\cite{mpii} + AP10K~\cite{ap10k} + APT36K~\cite{apt36k} + WholeBody~\cite{wholebody} & 76.99 (1.24 $\uparrow$) & 60.74 (1.24 $\uparrow$) & 78.90 (0.35 $\uparrow$) \\
    \midrule
    \end{tabular}
    \label{tab:pretrain}
    }
\end{table}

\paragraph{Impact of Input Resolutions.}
It has been shown that training on high-resolution images leads to the acquisition of more generalized and robust features, thereby improving performance under corruptions~\cite{robodepth}. However, our investigation, detailed in Table~\ref{tab:input_size}, examining two input sizes (256$\times$192 and 384$\times$288) across four models (\textit{i.e.}, SimpleBaseline~\cite{simplebaseline} with the Res50 and Res152 backbones and HRNet~\cite{hrnet} with the HRNet-W32 and HRNet-W48 backbones), reveals subtle distinctions. While larger input resolutions indeed enhance mean average precision (mAP) on both clean and corrupted images, the relative improvement under corruptions is not as pronounced as on clean images, leading to a lower mean relative rank (mRR) value. Consequently, increasing input resolution enhances model performance on clean images but also increases susceptibility to corruptions.

\paragraph{Impact of Post-processing.}
We investigate the impact of two commonly used post-processing techniques for top-down methods, UDP~\cite{udp} and DARK~\cite{dark}. The results are presented in Table~\ref{tab:post_process}. We evaluate two models, HRNet W32 and HRNet W48, trained on COCO~\cite{coco} and tested on the COCO-C dataset, both with and without post-processing. The findings indicate that while post-processing techniques effectively improve the mAP on both clean and corrupted images, there is a slight degradation in corruption robustness as measured by mRR values.

\paragraph{Impact of Training on Multiple Datasets.}
Training on diverse datasets has long been acknowledged as an effective strategy for enhancing model generalization and robustness. Exposure to a wider range of data during training enables models to learn more generalized features, thereby enhancing performance on unseen data. To investigate the impact of pre-training datasets on corruption robustness, we trained ViTPose (ViT-B)~\cite{vitpose} on various dataset combinations, including COCO~\cite{coco}, AI Challenger (AIC)~\cite{aic}, MPII~\cite{mpii}, CrowdPose~\cite{crowdpose}, AP10K~\cite{ap10k}, APT36K~\cite{apt36k}, and WholeBody~\cite{wholebody}.

Table~\ref{tab:pretrain} shows that training on multiple datasets enhances both mAP and mRR under corruptions compared to using a single COCO dataset (Row 1). However, we observed that more datasets do not always yield better results: the type and distribution of data also play a crucial role. For instance, when comparing Row 3 and Row 4, we found that crowd pose data (CrowdPose~\cite{crowdpose}) significantly improved mAP on both clean and corrupted images more than animal pose data (AP10K~\cite{ap10k}) and whole-body pose data (WholeBody~\cite{wholebody}). However, the latter achieved a higher mRR, indicating a greater improvement in robustness against corruptions relative to clean images.

\begin{table*}[t!]
    \caption{
    Performance evaluation of ViTPose (ViT-B) on COCO-C dataset with varied data augmentations. A: blur and noise, B: compression and color alteration, C: lighting adjustments, and D: occlusion with dropout.
    }
    \centering\scalebox{0.6}{
    \begin{tabular}{r|cc|ccc|ccc|ccc|ccc|ccc}
    \toprule
    \multirow{2}{*}{\textbf{Aug Set}} & \multicolumn{2}{c}{\textbf{Clean}}\vline & \multicolumn{3}{c}{\textbf{Overall Robustness}}\vline & \multicolumn{3}{c}{\textbf{Blur \& Noise}}\vline & \multicolumn{3}{c}{\textbf{Compression \& Color}}\vline & \multicolumn{3}{c}{\textbf{Lightning}}\vline & \multicolumn{3}{c}{\textbf{Mask}}
    \\
    ~ & \textcolor{robo_blue}{\textbf{mAP}} & \textcolor{robo_red}{\textbf{mAR}} & \textcolor{robo_blue}{\textbf{mAP}} & \textcolor{robo_red}{\textbf{mAR}} & \textcolor{robo_green}{\textbf{mRR}} & \textcolor{robo_blue}{\textbf{mAP}} & \textcolor{robo_red}{\textbf{mAR}} & \textcolor{robo_green}{\textbf{mRR}} & \textcolor{robo_blue}{\textbf{mAP}} & \textcolor{robo_red}{\textbf{mAR}} & \textcolor{robo_green}{\textbf{mRR}} & \textcolor{robo_blue}{\textbf{mAP}} & \textcolor{robo_red}{\textbf{mAR}} & \textcolor{robo_green}{\textbf{mRR}} & \textcolor{robo_blue}{\textbf{mAP}} & \textcolor{robo_red}{\textbf{mAR}} & \textcolor{robo_green}{\textbf{mRR}} \\
    \midrule\midrule
    - & 75.75 & 80.99 & 59.50 & 65.13 & 78.55 & 52.45 & 58.38 & 69.25 & \cellcolor{blue!10} 63.0 & 68.32 & \cellcolor{green!10} 83.17 & 61.91 & 67.53 & 81.73 & 62.90 & \cellcolor{red!10} 68.62 & 83.04\\ 
    \midrule
    A & 75.81 & 81.08 & 60.16 & 65.79 & 79.35 & 53.76 & 59.67 & 70.91 & \cellcolor{blue!10} 63.53 & 68.86 & \cellcolor{green!10} 83.80 & 62.16 & 67.82 & 82.00 & 63.20 & \cellcolor{red!10} 68.91 & 83.37\\ 
    B & \textbf{76.10} & \textbf{81.35} & 60.75 & 66.33 & 79.83 & 53.89 & 59.81 & 70.82 & 63.94 & 69.20 & 84.03 & 63.26 & 68.82 & 83.14 & \cellcolor{blue!10} 64.18 & \cellcolor{red!10} 69.84 & \cellcolor{green!10} 84.34\\ 
    C & 75.83 & 81.02 & 61.01 & 66.53 & 80.46 & 53.74 & 59.56 & 70.87 & 64.11 & 69.35 & 84.54 & \cellcolor{blue!10} 64.17 & 69.60 & \cellcolor{green!10} 84.62 & 64.06 & \cellcolor{red!10} 69.78 & 84.48\\ 
    D & 75.79 & 81.10 & 60.51 & 66.15 & 79.84 & 53.07 & 59.01 & 70.03 & 63.61 & 68.98 & 83.94 & 62.68 & 68.31 & 82.71 & \cellcolor{blue!10} 66.98 & \cellcolor{red!10} 72.63 & \cellcolor{green!10} 88.38\\ 
    AB & 76.01 & 81.26 & 61.24 & 66.76 & 80.56 & 54.86 & 60.59 & 72.17 & \cellcolor{blue!10} 64.34 & 69.56 & \cellcolor{green!10} 84.65 & 63.60 & 69.15 & 83.67 & 63.99 & \cellcolor{red!10} 69.66 & 84.18\\ 
    ABC & 76.00 & 81.22 & 62.05 & 67.55 & 81.65 & 55.25 & 60.99 & 72.69 & 64.82 & 70.07 & 85.28 & \cellcolor{blue!10} 65.28 & \cellcolor{red!10} 70.71 & \cellcolor{green!10} 85.89 & 64.52 & 70.23 & 84.89\\ 
    ABCD & 75.76 & 81.05 & \textbf{62.49} & \textbf{67.97} & \textbf{82.48} & \textbf{55.46} & \textbf{61.13} & \textbf{73.20} & \textbf{64.97} & \textbf{70.25} & \textbf{85.76} & \textbf{65.41} & \textbf{70.85} & \textbf{86.34} & \cellcolor{blue!10} \textbf{67.32} & \cellcolor{red!10} \textbf{72.96} & \cellcolor{green!10} \textbf{88.86}\\ 
    \bottomrule
    \end{tabular}
    }
\label{tab:train_on_aug}
\end{table*}

\paragraph{Impact of Data Augmentation.}
We investigated the impact of data augmentation during training to enhance robustness against diverse natural corruptions. Utilizing the state-of-the-art ViTPose model (ViT-B) as our base, we employed four augmentation sets: blur and noise (A), compression and color alteration (B), lighting adjustments (C), and occlusion with dropout (D). Detailed configurations for each set are available in the Appendix.

Initially, we individually applied each augmentation set. As illustrated in Table~\ref{tab:train_on_aug}, each augmentation set notably improved robustness metrics against corresponding corruptions. For instance, Corrupted mAP and mRR on motion blur increased from 59.50 and 78.55 to 60.16 and 79.35, respectively, with the application of blur and noise augmentations during training.
Furthermore, by progressively combining more augmentation techniques (from A to a combination of A, B, C, and D), performance substantially surpassed the baseline, showcasing the cumulative effect of multiple augmentations on robustness.

\section{Conclusion}\label{sec:conclusion}
In this study, we performed a thorough robustness benchmark for pose estimation models across diverse real-world corruptions. We assessed 60 models derived from 15 leading pose estimation methods, covering human and animal bodies across three representative datasets: COCO-C, OCHuman-C, and AP10K-C datasets. Our findings offer crucial insights into model reliability under corruptions and highlight key factors that can improve robustness and generalization. We hope this study could facilitate the development of more resilient and effective pose models and training methodologies.

{
    \small
    \bibliographystyle{plain}
    \bibliography{main}
}

\newpage
\appendix
\section*{Appendix}

This section outlines the details of corruption types and augmentations in Section~\ref{sec:appendix_corr} and Section~\ref{sec:appendix_aug}, presents additional experimental results in Section~\ref{sec:appendix_experiment}, and discusses the limitations and future work in Section~\ref{sec:appendix_disscussion}.

\section{Definition of Corruption and Severity Levels}\label{sec:appendix_corr}
\begin{table}[]
    \centering
    \caption{
    Definitions of corruption and parameters used for each severity level.}
    \scalebox{0.85}{
    \begin{tabular}{l|c|cccccc}
    \toprule
    \textbf{Type} & \textbf{Parameters} & \textbf{Level 1} & \textbf{Level 2} & \textbf{Level 3} & \textbf{Level 4} & \textbf{Level 5}  \\
    \midrule\midrule
    Motion Blur & (radius, sigma) & (10, 3) & (15, 5) & (15, 8) & (15, 12) & (20, 15)\\
    Gaussian Noise & sigma & 1 & 2 & 3 & 4 & 6 \\
    \multirow{2}{*}{Impulse Noise} & proportion of image pixels & \multirow{2}{*}{3} & \multirow{2}{*}{6} & \multirow{2}{*}{9} & \multirow{2}{*}{17} & \multirow{2}{*}{27} \\
    ~ & to replace with noise (\%) & \\
    \midrule
    \multirow{2}{*}{Pixelate} & proportion of image & \multirow{2}{*}{60} & \multirow{2}{*}{50} & \multirow{2}{*}{40} & \multirow{2}{*}{30} & \multirow{2}{*}{25}\\
    ~ & shape to resize (\%) & \\
    JPEG Compression & JPEG quality (\%) & 25 & 18 & 15 & 10 & 7 \\
    \multirow{2}{*}{Color Quant} & the number of bits to & \multirow{2}{*}{5} & \multirow{2}{*}{4} & \multirow{2}{*}{3} & \multirow{2}{*}{2} & \multirow{2}{*}{1}  \\
    ~ & keep for each channel \\
    \midrule
    \multirow{2}{*}{Brightness} & adjustment of value & \multirow{2}{*}{0.1} & \multirow{2}{*}{0.2} & \multirow{2}{*}{0.3} & \multirow{2}{*}{0.4} & \multirow{2}{*}{0.5} \\
    ~ & in HSV space \\
    Darkness & adjustment of gamma & 0.6 & 0.5 & 0.4 & 0.3 & 0.2\\
    Contrast & adjustment of std & 0.4 & 0.3 & 0.2 & 0.1 & 0.05 \\
    \midrule
    \multirow{2}{*}{Mask} & mask size for COCO-C, & \multirow{2}{*}{5, 20, 20} & \multirow{2}{*}{10, 25, 25} & \multirow{2}{*}{15, 30, 30} & \multirow{2}{*}{20, 35, 35} & \multirow{2}{*}{25, 40, 40}\\
    ~ &  OCHuman-C, AP10K-C & ~ & ~ & ~ & ~ & ~ \\
    \bottomrule
    \end{tabular}
    }
    \label{tab:appendix_corr_def}
\end{table}

The severity levels for 10 types of corruptions are detailed in Table~\ref{tab:appendix_corr_def}. These corruptions include:
\begin{itemize}
    \item Motion Blur: 
    is caused by the fast movement of cameras or subjects, resulting in a blurry appearance along the direction of motion. This is achieved by using a motion kernel with varying radius and sigma.

    \item Gaussian Noise: 
    is implemented by adding random values from a Gaussian distribution (zero mean, varying sigma) to image pixels, creating different noise levels.

    \item Impulse Noise: 
    also known as salt-and-pepper noise, is characterized by sporadic white and black pixels. This is achieved by randomly replacing a percentage of pixels with either black (0) or white (255) values.

    \item Pixelate: 
    reduces image quality by grouping blocks of pixels and assigning a single color value to each block, resulting in a blocky appearance. This is achieved by resizing the image to a lower resolution and then scaling it back up using nearest-neighbor interpolation.

    \item JPEG Compression: 
    reduces file size by discarding certain data components, leading to compression artifacts. This is achieved by saving the image in JPEG format with a quality setting below 30.

    \item Color Quant: 
    reduces the number of distinct colors in an image, creating a posterized effect. This is achieved by reducing the bit depth of each color channel.

    \item Brightness: 
    uniformly increases the intensity of all pixels. It is implemented in the HSV space by a constant factor while ensuring values remain within the range of 0 to 255.

    \item Darkness: 
    uniformly decreases the intensity of all pixels. This is achieved by scaling down pixel values by a constant factor (gamma) in a normalized space.

    \item Contrast: 
    alters the difference in luminance or color to enhance object visibility. It is implemented by scaling the variance of pixel values by a constant factor, while maintaining the same mean and clipping values to stay within the range of 0 to 255.

    \item Mask:
    occludes parts of the image with black (255), simulating occlusion or missing data.

\end{itemize}

\section{Augmentation Definition}\label{sec:appendix_aug}
\begin{table}[]
    \caption{Definition of augmentation sets. }
    \centering
    \scalebox{1}{
    \begin{tabular}{l|l}
    \toprule
    \textbf{Augmentation set}  & \textbf{Augmentation Types} \\
    \midrule\midrule
    \multirow{2}{*}{A: blur and noise} & blur, median blur, Gaussian blur, \\
    ~ & Gaussian noise, ISO noise, motion blur \\
    \midrule
    \multirow{2}{*}{B: compression and color alteration} & color jitter, image compression, RGB shift,  \\
    ~ & to gray, pixel dropout, superpixels \\
    \midrule
    \multirow{4}{*}{C: lighting adjustments} & random HSV, random brightness, \\
    ~ & random contrast, random gamma, \\
    ~ & random sun flare, random shadow, \\
    ~ & rining overshoot\\
    \midrule
    D: occlusion with dropout & XY masking, grid dropout, coarse dropout\\
    \bottomrule
    \end{tabular}
    }
    \label{tab:appendix_aug_def}
\end{table}

Table~\ref{tab:appendix_aug_def} provides details of the augmentation types included in each set used in Section~\ref{sec:robust_enhance}.

\section{Experiment Results}\label{sec:appendix_experiment}
\begin{table}[]
    \caption{
    The mAP scores of various models on the COCO-C dataset. Scores are presented from left to right for clean images, all corrupted images, and images corrupted by: Motion Blur, Gaussian Noise, Impulse Noise, Pixelate, JPEG Compression, Color Quantization, Brightness, Darkness, Contrast, and Mask, respectively. 
    }
    \centering\scalebox{0.64}{
    \begin{tabular}{rc|cc|cccccccccc}
    \toprule
    \textbf{Method} & \textbf{Backbone} & \textbf{Clean} & \textbf{Corr} & \textbf{Motion} & \textbf{Gauss} & \textbf{Impul} & \textbf{Pixel} & \textbf{JPEG} & \textbf{Quant} & \textbf{Bright} & \textbf{Dark} & \textbf{Contr} & \textbf{Mask} \\
    \midrule\midrule
    Hourglass~\cite{hourglass} & Hourglass-52 & 72.58 & 52.86 & 40.64 & 45.18 & 45.76 & 58.00 & 54.63 & 58.39 & 67.88 & 49.55 & 49.28 & 59.34\\
    SimpleBaseline~\cite{simplebaseline} & Res50 & 71.82 & 52.32 & 39.18 & 46.92 & 47.27 & 56.21 & 53.23 & 57.71 & 66.71 & 48.61 & 48.30 & 59.05\\ 
    SimpleBaseline~\cite{simplebaseline} & Res101 & 72.78 & 53.68 & 40.05 & 48.92 & 48.40 & 57.95 & 55.01 & 59.37 & 68.09 & 49.40 & 49.51 & 60.10\\ 
    SimpleBaseline~\cite{simplebaseline} & Res152 & 73.59 & 54.84 & 41.14 & 50.26 & 50.28 & 58.67 & 56.29 & 60.49 & 69.08 & 50.06 & 51.14 & 61.03\\ 
    HRNet~\cite{hrnet} & HRNet-W32 & 74.90 & 56.33 & 43.27 & 50.86 & 52.13 & 58.46 & 56.51 & 61.16 & 70.65 & 52.74 & 54.86 & 62.61\\ 
    HRNet~\cite{hrnet} & HRNet-W48 & 75.58 & 57.52 & 43.73 & 53.05 & 54.10 & 59.94 & 57.73 & 62.42 & 71.62 & 52.61 & 56.14 & 63.85\\ 
    MSPN~\cite{mspn} & MSPN-50 & 72.28 & 52.37 & 39.81 & 45.44 & 45.46 & 57.74 & 52.60 & 58.05 & 67.19 & 49.58 & 48.67 & 59.17\\ 
    MSPN~\cite{mspn} & 2xMSPN-50 & 75.35 & 56.71 & 43.88 & 50.89 & 51.01 & 61.69 & 56.80 & 62.09 & 71.18 & 52.21 & 54.23 & 63.11\\ 
    MSPN~\cite{mspn} & 3xMSPN-50 & 75.85 & 58.12 & 44.48 & 53.36 & 53.71 & 63.34 & 58.89 & 63.32 & 72.03 & 53.35 & 55.65 & 63.12\\ 
    MSPN~\cite{mspn} & 4xMSPN-50 & 76.49 & 58.83 & 44.76 & 54.66 & 54.84 & 64.52 & 59.75 & 63.90 & 72.60 & 53.64 & 55.63 & 64.03\\ 
    RSN~\cite{rsn} & RSN-18 & 70.40 & 51.70 & 39.96 & 45.51 & 45.49 & 57.54 & 52.25 & 56.77 & 65.65 & 48.70 & 48.18 & 56.94\\ 
    RSN~\cite{rsn} & RSN-50 & 72.41 & 54.36 & 42.02 & 49.51 & 49.57 & 60.11 & 54.35 & 59.21 & 67.87 & 50.61 & 51.29 & 59.09\\ 
    RSN~\cite{rsn} & 2xRSN-50 & 74.76 & 57.70 & 44.51 & 53.22 & 53.30 & 64.25 & 58.51 & 62.50 & 70.82 & 53.42 & 54.56 & 61.91\\ 
    RSN~\cite{rsn} & 3xRSN-50 & 75.05 & 58.19 & 44.69 & 54.11 & 54.03 & 64.96 & 60.03 & 63.20 & 71.41 & 53.39 & 53.69 & 62.42\\ 
    TransPose~\cite{transpose} & Res50-A3 & 71.54 & 51.54 & 37.85 & 45.02 & 44.84 & 56.38 & 52.81 & 57.20 & 66.43 & 48.59 & 48.12 & 58.17\\ 
    TransPose~\cite{transpose} & Res50-A4 & 72.64 & 52.67 & 38.68 & 45.79 & 46.02 & 57.31 & 54.55 & 58.42 & 67.65 & 49.61 & 49.45 & 59.25\\ 
    TransPose~\cite{transpose} & HRNet-W32 & 74.17 & 54.93 & 40.64 & 48.68 & 49.96 & 58.55 & 55.24 & 60.12 & 69.37 & 52.36 & 53.14 & 61.24\\ 
    TransPose~\cite{transpose} & HRNet-W48-A4 & 75.28 & 56.54 & 41.93 & 50.55 & 51.34 & 59.78 & 56.36 & 61.77 & 71.10 & 53.90 & 55.71 & 63.00\\ 
    TransPose~\cite{transpose} & HRNet-W48-A6 & 75.78 & 56.90 & 41.58 & 51.38 & 52.32 & 60.18 & 57.07 & 62.07 & 71.54 & 53.26 & 55.97 & 63.67\\ 
    HRFormer~\cite{hrformer} & HRFormer-S & 73.84 & 54.87 & 42.03 & 49.40 & 50.47 & 57.05 & 54.38 & 59.97 & 69.27 & 51.50 & 53.08 & 61.54\\ 
    HRFormer~\cite{hrformer} & HRFormer-B & 75.37 & 56.67 & 43.08 & 52.07 & 52.87 & 58.19 & 56.53 & 62.41 & 71.38 & 51.72 & 54.92 & 63.55\\ 
    LiteHRNet~\cite{litehrnet} & LiteHRNet-18 & 64.16 & 45.80 & 36.52 & 37.04 & 37.90 & 53.39 & 49.05 & 51.17 & 59.22 & 42.48 & 41.00 & 50.20\\ 
    LiteHRNet~\cite{litehrnet} & LiteHRNet-30 & 67.54 & 49.08 & 38.48 & 40.42 & 42.00 & 55.56 & 51.95 & 54.19 & 63.06 & 45.57 & 46.08 & 53.45\\ 
    ViTPose~\cite{vitpose} & ViT-S & 73.92 & 56.73 & 41.80 & 53.35 & 53.59 & 61.21 & 58.60 & 61.33 & 70.05 & 52.61 & 54.36 & 60.43\\ 
    ViTPose~\cite{vitpose} & ViT-B & 75.75 & 59.50 & 43.29 & 56.97 & 57.09 & 63.75 & 61.30 & 63.95 & 72.33 & 55.03 & 58.38 & 62.90\\ 
    ViTPose~\cite{vitpose} & ViT-L & 78.18 & 63.83 & 45.70 & 62.95 & 63.67 & 67.84 & 65.49 & 68.22 & 75.55 & 59.02 & 63.46 & 66.43\\ 
    ViTPose~\cite{vitpose} & ViT-H & 78.84 & 65.02 & 46.43 & 64.71 & 65.53 & 69.19 & 66.97 & 69.52 & 76.55 & 59.81 & 64.52 & 66.93\\ 
    SimCC~\cite{simcc} & Res50 & 72.07 & 52.25 & 38.65 & 46.67 & 46.60 & 56.01 & 53.28 & 58.23 & 67.16 & 48.43 & 48.25 & 59.17\\ 
    SimCC~\cite{simcc} & MobileNetV2 & 61.98 & 41.45 & 31.58 & 33.09 & 33.53 & 46.49 & 43.73 & 46.95 & 55.75 & 38.40 & 36.07 & 48.89\\ 
    SimCC~\cite{simcc} & vipnas-mbv3 & 69.48 & 50.31 & 38.29 & 43.35 & 44.54 & 54.82 & 51.40 & 55.47 & 64.63 & 46.94 & 47.43 & 56.23\\ 
    \midrule
    AE~\cite{ae} & HRNet-W32 & 65.55 & 42.53 & 33.68 & 35.82 & 36.90 & 40.97 & 34.51 & 48.99 & 59.22 & 39.51 & 42.90 & 52.82\\ 
    HigherHRNet~\cite{higherhrnet} & HRNet-W32 & 67.66 & 43.71 & 34.35 & 36.12 & 37.11 & 42.64 & 35.94 & 51.07 & 61.07 & 40.61 & 44.23 & 53.95\\ 
    HigherHRNet~\cite{higherhrnet} & HRNet-W48 & 68.51 & 44.91 & 35.05 & 39.10 & 39.96 & 42.49 & 36.96 & 51.91 & 62.04 & 41.08 & 45.48 & 55.00\\ 
    DEKR~\cite{dekr} & HRNet-W32 & 68.64 & 45.01 & 33.49 & 37.56 & 38.96 & 44.85 & 39.00 & 52.34 & 62.62 & 41.52 & 45.19 & 54.57\\ 
    DEKR~\cite{dekr} & HRNet-W48 & 71.37 & 46.30 & 33.79 & 40.80 & 42.14 & 41.79 & 38.34 & 54.25 & 65.03 & 43.00 & 47.14 & 56.77\\ 
    \midrule
    PRTR~\cite{prtr} & Res50 & 67.34 & 45.93 & 33.72 & 40.15 & 40.26 & 45.88 & 45.47 & 52.10 & 61.39 & 42.93 & 43.08 & 54.29\\ 
    PRTR~\cite{prtr} & Res101 & 68.95 & 48.02 & 34.73 & 43.44 & 43.56 & 48.62 & 46.64 & 53.93 & 63.19 & 43.98 & 46.16 & 55.98\\ 
    PRTR~\cite{prtr} & HRNet-W32 & 71.71 & 51.21 & 38.20 & 45.58 & 47.32 & 52.91 & 49.54 & 56.73 & 66.43 & 47.27 & 49.41 & 58.75\\ 
    Poseur~\cite{poseur} & MobileNetV2 & 70.55 & 49.31 & 38.18 & 39.82 & 40.91 & 54.77 & 51.61 & 55.53 & 64.88 & 46.24 & 44.54 & 56.56\\ 
    Poseur~\cite{poseur} & Res50 & 74.22 & 53.79 & 40.57 & 47.87 & 48.02 & 56.62 & 54.15 & 59.46 & 68.97 & 50.57 & 50.26 & 61.40\\ 
    Poseur~\cite{poseur} & HRNet-W32 & 75.52 & 55.91 & 42.93 & 48.98 & 50.45 & 57.62 & 55.72 & 61.17 & 71.17 & 52.75 & 54.77 & 63.57\\ 
    Poseur~\cite{poseur} & HRNet-W48 & 77.62 & 59.97 & 45.07 & 55.51 & 56.34 & 59.15 & 56.95 & 65.73 & 74.43 & 57.89 & 60.35 & 68.26\\ 
    Poseur~\cite{poseur} & HRFormer-T & 73.26 & 51.92 & 39.88 & 43.39 & 44.42 & 54.45 & 51.52 & 57.70 & 67.79 & 49.82 & 50.31 & 59.95\\ 
    Poseur~\cite{poseur} & HRFormer-S & 75.49 & 55.95 & 42.65 & 49.85 & 50.59 & 57.90 & 54.98 & 61.74 & 71.06 & 52.81 & 54.53 & 63.40\\ 
    Poseur~\cite{poseur} & HRFormer-B & 77.97 & 61.84 & 46.08 & 58.52 & 58.84 & 62.40 & 60.02 & 67.38 & 75.15 & 59.37 & 62.11 & 68.53\\ 
    Poseur~\cite{poseur} & ViT-B & 76.72 & 62.17 & 45.96 & 58.92 & 58.97 & 64.39 & 63.13 & 66.98 & 73.89 & 59.56 & 61.89 & 67.99\\ 
    \bottomrule
    \end{tabular}
    }
    \label{tab:supp_coco_map}
\end{table}

\begin{table}[]
    \caption{
    The mAR scores of various models on the COCO-C dataset. Scores are presented from left to right for clean images, all corrupted images, and images corrupted by: Motion Blur, Gaussian Noise, Impulse Noise, Pixelate, JPEG Compression, Color Quantization, Brightness, Darkness, Contrast, and Mask, respectively. 
    }
    \centering\scalebox{0.64}{
    \begin{tabular}{rc|cc|cccccccccc}
    \toprule
    \textbf{Method} & \textbf{Backbone} & \textbf{Clean} & \textbf{Corr} & \textbf{Motion} & \textbf{Gauss} & \textbf{Impul} & \textbf{Pixel} & \textbf{JPEG} & \textbf{Quant} & \textbf{Bright} & \textbf{Dark} & \textbf{Contr} & \textbf{Mask} \\
    \midrule\midrule
    Hourglass~\cite{hourglass} & Hourglass-52 & 77.98 & 58.49 & 46.10 & 50.96 & 51.59 & 63.33 & 60.05 & 63.96 & 73.29 & 55.23 & 55.31 & 65.06\\
    SimpleBaseline~\cite{simplebaseline} & Res50 & 77.38 & 58.18 & 45.13 & 53.12 & 53.46 & 61.61 & 58.81 & 63.43 & 72.35 & 54.65 & 54.42 & 64.84\\ 
    SimpleBaseline~\cite{simplebaseline} & Res101 & 78.28 & 59.51 & 46.00 & 55.02 & 54.57 & 63.30 & 60.46 & 65.03 & 73.64 & 55.42 & 55.71 & 65.93\\ 
    SimpleBaseline~\cite{simplebaseline} & Res152 & 79.09 & 60.63 & 47.05 & 56.32 & 56.27 & 64.04 & 61.68 & 66.15 & 74.69 & 55.88 & 57.31 & 66.95\\ 
    HRNet~\cite{hrnet} & HRNet-W32 & 80.38 & 61.93 & 48.78 & 56.82 & 58.14 & 63.70 & 61.88 & 66.64 & 76.01 & 58.48 & 60.52 & 68.29\\ 
    HRNet~\cite{hrnet} & HRNet-W48 & 80.85 & 63.04 & 49.24 & 58.95 & 60.07 & 65.02 & 63.04 & 67.89 & 76.87 & 58.27 & 61.65 & 69.42\\ 
    MSPN~\cite{mspn} & MSPN-50 & 78.80 & 59.20 & 46.07 & 52.65 & 52.72 & 64.29 & 59.29 & 64.91 & 73.85 & 56.69 & 55.64 & 65.89\\ 
    MSPN~\cite{mspn} & 2xMSPN-50 & 81.56 & 63.32 & 50.08 & 57.82 & 57.99 & 67.85 & 63.33 & 68.72 & 77.56 & 59.13 & 60.98 & 69.72\\ 
    MSPN~\cite{mspn} & 3xMSPN-50 & 82.06 & 64.81 & 50.83 & 60.36 & 60.84 & 69.50 & 65.51 & 69.99 & 78.35 & 60.19 & 62.65 & 69.85\\ 
    MSPN~\cite{mspn} & 4xMSPN-50 & 82.62 & 65.59 & 51.23 & 61.93 & 62.03 & 70.71 & 66.32 & 70.57 & 78.92 & 60.69 & 62.64 & 70.83\\ 
    RSN~\cite{rsn} & RSN-18 & 77.28 & 58.74 & 46.26 & 53.01 & 52.99 & 64.25 & 59.12 & 63.86 & 72.54 & 56.01 & 55.39 & 64.02\\ 
    RSN~\cite{rsn} & RSN-50 & 79.07 & 61.41 & 48.50 & 57.15 & 57.24 & 66.65 & 61.23 & 66.17 & 74.66 & 57.88 & 58.55 & 66.09\\ 
    RSN~\cite{rsn} & 2xRSN-50 & 81.02 & 64.45 & 50.99 & 60.45 & 60.54 & 70.47 & 65.10 & 69.10 & 77.16 & 60.38 & 61.61 & 68.69\\ 
    RSN~\cite{rsn} & 3xRSN-50 & 81.37 & 64.91 & 51.14 & 61.19 & 61.20 & 71.28 & 66.52 & 69.82 & 77.73 & 60.33 & 60.69 & 69.19\\ 
    TransPose~\cite{transpose} & Res50-A3 & 77.03 & 57.19 & 43.30 & 50.94 & 50.76 & 61.72 & 58.17 & 62.71 & 71.89 & 54.45 & 54.11 & 63.89\\ 
    TransPose~\cite{transpose} & Res50-A4 & 77.96 & 58.26 & 44.19 & 51.67 & 51.89 & 62.50 & 59.92 & 63.86 & 72.97 & 55.39 & 55.36 & 64.85\\ 
    TransPose~\cite{transpose} & HRNet-W32 & 79.45 & 60.44 & 46.02 & 54.55 & 55.77 & 63.63 & 60.51 & 65.63 & 74.61 & 58.07 & 58.78 & 66.79\\ 
    TransPose~\cite{transpose} & HRNet-W48-A4 & 80.33 & 61.96 & 47.33 & 56.36 & 57.13 & 64.83 & 61.60 & 67.12 & 76.17 & 59.44 & 61.13 & 68.50\\ 
    TransPose~\cite{transpose} & HRNet-W48-A6 & 80.79 & 62.47 & 47.20 & 57.48 & 58.40 & 65.30 & 62.42 & 67.52 & 76.61 & 58.97 & 61.66 & 69.14\\ 
    HRFormer~\cite{hrformer} & HRFormer-S & 79.27 & 60.67 & 47.70 & 55.68 & 56.76 & 62.38 & 59.97 & 65.70 & 74.81 & 57.49 & 59.02 & 67.21\\ 
    HRFormer~\cite{hrformer} & HRFormer-B & 80.67 & 62.31 & 48.82 & 58.15 & 58.88 & 63.48 & 61.95 & 68.00 & 76.67 & 57.45 & 60.51 & 69.22\\ 
    LiteHRNet~\cite{litehrnet} & LiteHRNet-18 & 70.45 & 52.11 & 42.55 & 43.62 & 44.57 & 59.32 & 55.10 & 57.42 & 65.51 & 48.97 & 47.55 & 56.48\\ 
    LiteHRNet~\cite{litehrnet} & LiteHRNet-30 & 73.61 & 55.33 & 44.42 & 47.17 & 48.65 & 61.36 & 57.88 & 60.31 & 69.07 & 52.05 & 52.65 & 59.75\\ 
    ViTPose~\cite{vitpose} & ViT-S & 79.24 & 62.56 & 47.69 & 59.62 & 59.85 & 66.64 & 64.15 & 66.98 & 75.47 & 58.42 & 60.45 & 66.29\\ 
    ViTPose~\cite{vitpose} & ViT-B & 80.99 & 65.13 & 49.21 & 62.90 & 63.04 & 68.94 & 66.60 & 69.42 & 77.56 & 60.71 & 64.33 & 68.62\\ 
    ViTPose~\cite{vitpose} & ViT-L & 83.43 & 69.37 & 51.80 & 68.58 & 69.32 & 73.01 & 70.64 & 73.60 & 80.77 & 64.63 & 69.27 & 72.08\\ 
    ViTPose~\cite{vitpose} & ViT-H & 83.92 & 70.39 & 52.50 & 70.14 & 71.04 & 74.24 & 71.99 & 74.72 & 81.56 & 65.15 & 70.16 & 72.37\\ 
    SimCC~\cite{simcc} & Res50 & 78.16 & 58.36 & 45.00 & 52.83 & 52.78 & 61.80 & 59.10 & 64.18 & 73.14 & 54.71 & 54.49 & 65.53\\ 
    SimCC~\cite{simcc} & MobileNetV2 & 67.81 & 47.05 & 36.72 & 38.79 & 39.47 & 51.82 & 49.19 & 52.55 & 61.58 & 44.12 & 41.54 & 54.71\\ 
    SimCC~\cite{simcc} & vipnas-mbv3 & 75.52 & 56.50 & 44.22 & 49.97 & 51.16 & 60.54 & 57.33 & 61.55 & 70.65 & 53.22 & 53.84 & 62.54\\ 
    \midrule
    AE~\cite{ae} & HRNet-W32 & 71.12 & 47.56 & 39.56 & 40.95 & 41.92 & 45.64 & 38.43 & 53.96 & 64.74 & 44.45 & 47.42 & 58.53\\ 
    HigherHRNet~\cite{higherhrnet} & HRNet-W32 & 72.38 & 48.15 & 39.93 & 40.32 & 41.25 & 46.82 & 39.42 & 55.49 & 65.88 & 44.95 & 48.36 & 59.14\\ 
    HigherHRNet~\cite{higherhrnet} & HRNet-W48 & 73.07 & 49.17 & 40.36 & 43.65 & 44.46 & 46.08 & 40.28 & 55.94 & 66.55 & 45.16 & 49.32 & 59.86\\ 
    DEKR~\cite{dekr} & HRNet-W32 & 73.49 & 50.11 & 40.02 & 42.62 & 43.87 & 49.81 & 43.39 & 57.13 & 67.65 & 46.72 & 49.73 & 60.12\\ 
    DEKR~\cite{dekr} & HRNet-W48 & 76.23 & 51.46 & 40.33 & 46.18 & 47.33 & 46.41 & 42.73 & 58.97 & 70.03 & 48.33 & 51.73 & 62.57\\ 
    \midrule
    PRTR~\cite{prtr} & Res50 & 74.87 & 54.21 & 42.07 & 48.67 & 48.94 & 54.27 & 53.93 & 59.92 & 69.36 & 51.17 & 51.27 & 62.48 \\ 
    PRTR~\cite{prtr} & Res101 & 76.36 & 56.36 & 43.17 & 52.09 & 52.34 & 57.39 & 55.11 & 61.82 & 70.97 & 52.15 & 54.42 & 64.15\\ 
    PRTR~\cite{prtr} & HRNet-W32 & 78.30 & 58.65 & 45.77 & 53.29 & 55.02 & 60.58 & 57.25 & 63.75 & 73.39 & 54.61 & 56.68 & 66.12\\ 
    Poseur~\cite{poseur} & MobileNetV2 & 76.19 & 55.12 & 43.78 & 45.70 & 46.89 & 60.29 & 57.42 & 61.22 & 70.65 & 52.31 & 50.43 & 62.55\\ 
    Poseur~\cite{poseur} & Res50 & 79.59 & 59.57 & 46.40 & 53.88 & 54.06 & 62.07 & 59.78 & 65.01 & 74.57 & 56.58 & 56.08 & 67.28\\ 
    Poseur~\cite{poseur} & HRNet-W32 & 80.65 & 61.53 & 48.68 & 54.92 & 56.39 & 62.97 & 61.29 & 66.51 & 76.39 & 58.54 & 60.34 & 69.26\\ 
    Poseur~\cite{poseur} & HRNet-W48 & 82.33 & 65.03 & 50.54 & 60.70 & 61.57 & 64.12 & 62.07 & 70.52 & 79.17 & 62.95 & 65.22 & 73.41\\ 
    Poseur~\cite{poseur} & HRFormer-T & 78.71 & 57.72 & 45.63 & 49.43 & 50.43 & 60.05 & 57.22 & 63.24 & 73.36 & 55.88 & 56.04 & 65.90\\ 
    Poseur~\cite{poseur} & HRFormer-S & 80.78 & 61.79 & 48.71 & 56.03 & 56.87 & 63.46 & 60.74 & 67.29 & 76.48 & 58.80 & 60.29 & 69.27\\ 
    Poseur~\cite{poseur} & HRFormer-B & 82.95 & 67.23 & 52.03 & 64.13 & 64.37 & 67.56 & 65.40 & 72.61 & 80.11 & 64.78 & 67.29 & 74.05\\ 
    Poseur~\cite{poseur} & ViT-B & 81.92 & 67.54 & 51.75 & 64.47 & 64.50 & 69.40 & 68.29 & 72.18 & 79.00 & 65.04 & 67.28 & 73.52\\ 
    \bottomrule
    \end{tabular}
    }
    \label{tab:supp_coco_mar}
\end{table}

\begin{table}[]
    \caption{
    The mRR scores of various models on the COCO-C dataset. Scores are presented from left to right for clean images, all corrupted images, and images corrupted by: Motion Blur, Gaussian Noise, Impulse Noise, Pixelate, JPEG Compression, Color Quantization, Brightness, Darkness, Contrast, and Mask, respectively. }
    \centering\scalebox{0.64}{
    \begin{tabular}{rc|ccccccccccc}
    \toprule
    \textbf{Method} & \textbf{Backbone} & mRR (\%) & \textbf{Motion} & \textbf{Gauss} & \textbf{Impul} & \textbf{Pixel} & \textbf{JPEG} & \textbf{Quant} & \textbf{Bright} & \textbf{Dark} & \textbf{Contr} & \textbf{Mask} \\
    \midrule\midrule
    Hourglass~\cite{hourglass} & Hourglass-52 & 72.84 & 56.00 & 62.26 & 63.05 & 79.92 & 75.27 & 80.45 & 93.53 & 68.28 & 67.90 & 81.76\\
    SimpleBaseline~\cite{simplebaseline} & Res50 & 72.84 & 54.55 & 65.32 & 65.81 & 78.26 & 74.11 & 80.35 & 92.88 & 67.68 & 67.25 & 82.22\\ 
    SimpleBaseline~\cite{simplebaseline} & Res101 & 73.75 & 55.03 & 67.21 & 66.49 & 79.61 & 75.58 & 81.57 & 93.55 & 67.87 & 68.03 & 82.57\\ 
    SimpleBaseline~\cite{simplebaseline} & Res152 & 74.53 & 55.91 & 68.31 & 68.32 & 79.73 & 76.49 & 82.20 & 93.88 & 68.03 & 69.50 & 82.93\\ 
    HRNet~\cite{hrnet} & HRNet-W32 & 75.20 & 57.77 & 67.90 & 69.60 & 78.06 & 75.45 & 81.65 & 94.33 & 70.41 & 73.24 & 83.60\\ 
    HRNet~\cite{hrnet} & HRNet-W48 & 76.10 & 57.85 & 70.18 & 71.57 & 79.31 & 76.38 & 82.59 & 94.75 & 69.61 & 74.27 & 84.48\\ 
    MSPN~\cite{mspn} & MSPN-50 & 72.46 & 55.08 & 62.87 & 62.89 & 79.89 & 72.78 & 80.32 & 92.97 & 68.60 & 67.34 & 81.87\\ 
    MSPN~\cite{mspn} & 2xMSPN-50 & 75.26 & 58.24 & 67.53 & 67.69 & 81.86 & 75.37 & 82.40 & 94.46 & 69.29 & 71.97 & 83.76\\ 
    MSPN~\cite{mspn} & 3xMSPN-50 & 76.63 & 58.65 & 70.35 & 70.82 & 83.50 & 77.65 & 83.48 & 94.97 & 70.33 & 73.37 & 83.22\\ 
    MSPN~\cite{mspn} & 4xMSPN-50 & 76.92 & 58.51 & 71.46 & 71.69 & 84.35 & 78.11 & 83.54 & 94.92 & 70.13 & 72.72 & 83.70\\ 
    RSN~\cite{rsn} & RSN-18 & 73.44 & 56.76 & 64.65 & 64.61 & 81.73 & 74.22 & 80.64 & 93.26 & 69.18 & 68.45 & 80.89\\ 
    RSN~\cite{rsn} & RSN-50 & 75.08 & 58.03 & 68.37 & 68.46 & 83.02 & 75.06 & 81.77 & 93.73 & 69.90 & 70.84 & 81.61\\ 
    RSN~\cite{rsn} & 2xRSN-50 & 77.18 & 59.54 & 71.18 & 71.30 & 85.94 & 78.26 & 83.61 & 94.73 & 71.45 & 72.99 & 82.82\\ 
    RSN~\cite{rsn} & 3xRSN-50 & 77.54 & 59.55 & 72.10 & 71.99 & 86.56 & 80.00 & 84.22 & 95.15 & 71.14 & 71.55 & 83.18\\ 
    TransPose~\cite{transpose} & Res50-A3 & 72.04 & 52.91 & 62.92 & 62.68 & 78.81 & 73.81 & 79.95 & 92.85 & 67.92 & 67.26 & 81.30\\ 
    TransPose~\cite{transpose} & Res50-A4 & 72.52 & 53.25 & 63.04 & 63.35 & 78.90 & 75.10 & 80.43 & 93.14 & 68.30 & 68.08 & 81.57\\ 
    TransPose~\cite{transpose} & HRNet-W32 & 74.06 & 54.79 & 65.64 & 67.36 & 78.94 & 74.47 & 81.06 & 93.53 & 70.60 & 71.64 & 82.57\\ 
    TransPose~\cite{transpose} & HRNet-W48-A4 & 75.11 & 55.70 & 67.15 & 68.19 & 79.41 & 74.86 & 82.04 & 94.44 & 71.60 & 74.00 & 83.68\\ 
    TransPose~\cite{transpose} & HRNet-W48-A6 & 75.09 & 54.87 & 67.80 & 69.04 & 79.41 & 75.30 & 81.90 & 94.40 & 70.28 & 73.86 & 84.01\\ 
    HRFormer~\cite{hrformer} & HRFormer-S & 74.31 & 56.92 & 66.90 & 68.35 & 77.27 & 73.65 & 81.22 & 93.81 & 69.75 & 71.88 & 83.35\\ 
    HRFormer~\cite{hrformer} & HRFormer-B & 75.19 & 57.15 & 69.08 & 70.15 & 77.20 & 75.00 & 82.81 & 94.71 & 68.61 & 72.86 & 84.31\\ 
    LiteHRNet~\cite{litehrnet} & LiteHRNet-18 & 71.39 & 56.93 & 57.73 & 59.07 & 83.22 & 76.45 & 79.76 & 92.31 & 66.22 & 63.90 & 78.25\\ 
    LiteHRNet~\cite{litehrnet} & LiteHRNet-30 & 72.66 & 56.97 & 59.84 & 62.19 & 82.25 & 76.92 & 80.23 & 93.36 & 67.46 & 68.22 & 79.14\\\ 
    ViTPose~\cite{vitpose} & ViT-S & 76.75 & 56.56 & 72.17 & 72.49 & 82.81 & 79.28 & 82.97 & 94.77 & 71.17 & 73.54 & 81.75\\ 
    ViTPose~\cite{vitpose} & ViT-B & 78.55 & 57.15 & 75.22 & 75.37 & 84.16 & 80.92 & 84.42 & 95.48 & 72.65 & 77.07 & 83.04\\ 
    ViTPose~\cite{vitpose} & ViT-L & 81.65 & 58.45 & 80.52 & 81.44 & 86.78 & 83.76 & 87.27 & 96.64 & 75.49 & 81.17 & 84.97\\ 
    ViTPose~\cite{vitpose} & ViT-H & 82.46 & 58.89 & 82.08 & 83.11 & 87.76 & 84.94 & 88.18 & 97.09 & 75.86 & 81.84 & 84.89\\ 
    SimCC~\cite{simcc} & Res50 & 72.49 & 53.63 & 64.76 & 64.65 & 77.71 & 73.92 & 80.79 & 93.19 & 67.19 & 66.94 & 82.10\\ 
    SimCC~\cite{simcc} & MobileNetV2 & 66.88 & 50.96 & 53.40 & 54.10 & 75.01 & 70.56 & 75.76 & 89.94 & 61.96 & 58.20 & 78.88\\ 
    SimCC~\cite{simcc} & vipnas-mbv3 & 72.41 & 55.12 & 62.40 & 64.11 & 78.90 & 73.98 & 79.84 & 93.02 & 67.56 & 68.27 & 80.93\\ 
    \midrule
    AE~\cite{ae} & HRNet-W32 & 64.89 & 51.38 & 54.65 & 56.29 & 62.51 & 52.65 & 74.75 & 90.34 & 60.27 & 65.45 & 80.59\\ 
    HigherHRNet~\cite{higherhrnet} & HRNet-W32 & 64.60 & 50.77 & 53.38 & 54.84 & 63.01 & 53.11 & 75.47 & 90.26 & 60.01 & 65.36 & 79.74\\ 
    HigherHRNet~\cite{higherhrnet} & HRNet-W48 & 65.55 & 51.15 & 57.07 & 58.33 & 62.02 & 53.95 & 75.77 & 90.56 & 59.95 & 66.39 & 80.28\\ 
    DEKR~\cite{dekr} & HRNet-W32 & 65.57 & 48.79 & 54.72 & 56.76 & 65.34 & 56.81 & 76.25 & 91.23 & 60.49 & 65.84 & 79.49\\ 
    DEKR~\cite{dekr} & HRNet-W48 & 64.88 & 47.35 & 57.17 & 59.05 & 58.56 & 53.72 & 76.01 & 91.11 & 60.25 & 66.05 & 79.54\\ 
    \midrule
    PRTR~\cite{prtr} & Res50 & 68.20 & 50.08 & 59.63 & 59.79 & 68.14 & 67.52 & 77.36 & 91.17 & 63.75 & 63.98 & 80.62\\ 
    PRTR~\cite{prtr} & Res101 & 69.65 & 50.37 & 63.01 & 63.18 & 70.52 & 67.64 & 78.22 & 91.65 & 63.79 & 66.95 & 81.20\\ 
    PRTR~\cite{prtr} & HRNet-W32 & 71.42 & 53.27 & 63.56 & 65.98 & 73.79 & 69.09 & 79.11 & 92.63 & 65.92 & 68.90 & 81.93\\ 
    Poseur~\cite{poseur} & MobileNetV2 & 69.89 & 54.12 & 56.45 & 58.00 & 77.64 & 73.16 & 78.72 & 91.96 & 65.54 & 63.14 & 80.17\\ 
    Poseur~\cite{poseur} & Res50 & 72.47 & 54.67 & 64.50 & 64.70 & 76.29 & 72.96 & 80.12 & 92.92 & 68.14 & 67.71 & 82.72\\ 
    Poseur~\cite{poseur} & HRNet-W32 & 74.04 & 56.85 & 64.86 & 66.81 & 76.30 & 73.79 & 81.00 & 94.24 & 69.85 & 72.53 & 84.17\\ 
    Poseur~\cite{poseur} & HRNet-W48 & 77.26 & 58.07 & 71.51 & 72.59 & 76.21 & 73.38 & 84.69 & 95.90 & 74.59 & 77.75 & 87.95\\ 
    Poseur~\cite{poseur} & HRFormer-T & 70.87 & 54.43 & 59.22 & 60.63 & 74.32 & 70.32 & 78.76 & 92.53 & 68.01 & 68.67 & 81.83\\ 
    Poseur~\cite{poseur} & HRFormer-S & 74.12 & 56.50 & 66.03 & 67.02 & 76.70 & 72.83 & 81.79 & 94.14 & 69.95 & 72.24 & 83.99\\ 
    Poseur~\cite{poseur} & HRFormer-B & 79.32 & 59.10 & 75.06 & 75.47 & 80.04 & 76.98 & 86.43 & 96.38 & 76.15 & 79.67 & 87.90\\ 
    Poseur~\cite{poseur} & ViT-B & 81.03 & 59.90 & 76.80 & 76.86 & 83.93 & 82.29 & 87.30 & 96.31 & 77.64 & 80.68 & 88.63\\ 
    \bottomrule
    \end{tabular}
    }
    \label{tab:supp_coco_mrr}
\end{table}

\begin{table}[]
    \caption{
    The mAP scores of various models on the OCHuman-C dataset. Scores are presented from left to right for clean images, all corrupted images, and images corrupted by: Motion Blur, Gaussian Noise, Impulse Noise, Pixelate, JPEG Compression, Color Quantization, Brightness, Darkness, Contrast, and Mask, respectively.
    }
    \centering\scalebox{0.65}{
    \begin{tabular}{rc|cc|cccccccccc}
    \toprule
    \textbf{Method} & \textbf{Backbone} & \textbf{Clean} & \textbf{Corr} & \textbf{Motion} & \textbf{Gauss} & \textbf{Impul} & \textbf{Pixel} & \textbf{JPEG} & \textbf{Quant} & \textbf{Bright} & \textbf{Dark} & \textbf{Contr} & \textbf{Mask} \\
    \midrule\midrule
    SimpleBaseline~\cite{simplebaseline} & Res50 & 54.76 & 50.04 & 49.86 & 47.40 & 47.03 & 54.19 & 53.06 & 47.97 & 53.04 & 47.74 & 45.60 & 54.45\\ 
    SimpleBaseline~\cite{simplebaseline} & Res101 & 55.79 & 51.26 & 51.37 & 48.41 & 48.08 & 55.47 & 54.56 & 49.54 & 54.40 & 48.85 & 46.34 & 55.52\\ 
    SimpleBaseline~\cite{simplebaseline} & Res152 & 56.97 & 52.44 & 52.14 & 49.88 & 49.66 & 56.44 & 55.83 & 50.99 & 55.53 & 49.60 & 47.84 & 56.52\\ 
    ViTPose~\cite{vitpose} & ViT-S & 58.75 & 54.23 & 54.26 & 51.61 & 51.52 & 57.90 & 57.05 & 52.36 & 56.82 & 52.00 & 50.73 & 58.09\\ 
    ViTPose~\cite{vitpose} & ViT-B & 61.03 & 56.61 & 56.18 & 54.43 & 54.15 & 59.90 & 59.02 & 55.08 & 59.25 & 54.20 & 53.69 & 60.19\\ 
    ViTPose~\cite{vitpose} & ViT-L & 66.08 & 62.20 & 61.06 & 60.61 & 60.52 & 65.01 & 64.16 & 61.11 & 64.52 & 59.45 & 60.12 & 65.46\\ 
    ViTPose~\cite{vitpose} & ViT-H & 67.85 & 64.28 & 63.23 & 62.46 & 62.25 & 67.12 & 66.34 & 63.39 & 66.92 & 61.40 & 62.17 & 67.50\\
    \bottomrule
    \end{tabular}
    }
    \label{tab:supp_ochuman_map}
\end{table}

\begin{table}[]
    \caption{
    The mAR scores of various models on the OCHuman-C dataset. Scores are presented from left to right for clean images, all corrupted images, and images corrupted by: Motion Blur, Gaussian Noise, Impulse Noise, Pixelate, JPEG Compression, Color Quantization, Brightness, Darkness, Contrast, and Mask, respectively.
    }
    \centering\scalebox{0.65}{
    \begin{tabular}{rc|cc|cccccccccc}
    \toprule
    \textbf{Method} & \textbf{Backbone} & \textbf{Clean} & \textbf{Corr} & \textbf{Motion} & \textbf{Gauss} & \textbf{Impul} & \textbf{Pixel} & \textbf{JPEG} & \textbf{Quant} & \textbf{Bright} & \textbf{Dark} & \textbf{Contr} & \textbf{Mask} \\
    \midrule\midrule
    SimpleBaseline~\cite{simplebaseline} & Res50 & 59.30 & 55.03 & 54.81 & 52.77 & 52.47 & 58.76 & 57.84 & 52.97 & 57.83 & 52.95 & 50.95 & 59.01\\ 
    SimpleBaseline~\cite{simplebaseline} & Res101 & 60.64 & 56.33 & 56.32 & 53.80 & 53.52 & 60.15 & 59.30 & 54.63 & 59.39 & 54.10 & 51.71 & 60.33\\ 
    SimpleBaseline~\cite{simplebaseline} & Res152 & 61.53 & 57.30 & 56.88 & 55.09 & 54.91 & 60.92 & 60.32 & 56.00 & 60.09 & 54.69 & 53.06 & 61.03\\ 
    ViTPose~\cite{vitpose} & ViT-S & 62.92 & 58.82 & 58.67 & 56.52 & 56.52 & 62.07 & 61.36 & 57.01 & 61.27 & 56.94 & 55.53 & 62.28\\ 
    ViTPose~\cite{vitpose} & ViT-B & 65.12 & 61.15 & 60.47 & 59.17 & 58.93 & 64.17 & 63.36 & 59.66 & 63.57 & 59.07 & 58.52 & 64.58\\ 
    ViTPose~\cite{vitpose} & ViT-L & 69.48 & 66.09 & 64.89 & 64.67 & 64.52 & 68.63 & 67.77 & 65.21 & 68.23 & 63.60 & 64.24 & 69.20\\ 
    ViTPose~\cite{vitpose} & ViT-H & 71.04 & 67.74 & 66.66 & 66.17 & 65.87 & 70.26 & 69.63 & 67.04 & 70.07 & 65.28 & 65.85 & 70.61\\
    \bottomrule
    \end{tabular}
    }
    \label{tab:supp_ochuman_mar}
\end{table}

\begin{table}[]
    \caption{
    The mRR scores of various models on the OCHuman-C dataset. Scores are presented from left to right for clean images, all corrupted images, and images corrupted by: Motion Blur, Gaussian Noise, Impulse Noise, Pixelate, JPEG Compression, Color Quantization, Brightness, Darkness, Contrast, and Mask, respectively.
    }
    \centering\scalebox{0.65}{
    \begin{tabular}{rc|cc|ccccccccc}
    \toprule
    \textbf{Method} & \textbf{Backbone} & \textbf{mRR (\%)} & \textbf{Motion} & \textbf{Gauss} & \textbf{Impul} & \textbf{Pixel} & \textbf{JPEG} & \textbf{Quant} & \textbf{Bright} & \textbf{Dark} & \textbf{Contr} & \textbf{Mask} \\
    \midrule\midrule
    SimpleBaseline~\cite{simplebaseline} & Res50 & 91.37 & 91.06 & 86.57 & 85.89 & 98.96 & 96.89 & 87.59 & 96.87 & 87.18 & 83.28 & 99.44\\ 
    SimpleBaseline~\cite{simplebaseline} & Res101 & 91.87 & 92.08 & 86.77 & 86.19 & 99.42 & 97.79 & 88.80 & 97.51 & 87.57 & 83.06 & 99.52\\ 
    SimpleBaseline~\cite{simplebaseline} & Res152 & 92.05 & 91.52 & 87.55 & 87.16 & 99.07 & 98.00 & 89.50 & 97.46 & 87.06 & 83.97 & 99.20\\ 
    ViTPose~\cite{vitpose} & ViT-S & 92.31 & 92.36 & 87.85 & 87.69 & 98.56 & 97.10 & 89.12 & 96.71 & 88.52 & 86.35 & 98.87\\ 
    ViTPose~\cite{vitpose} & ViT-B & 92.75 & 92.05 & 89.17 & 88.73 & 98.15 & 96.71 & 90.24 & 97.08 & 88.80 & 87.97 & 98.62\\ 
    ViTPose~\cite{vitpose} & ViT-L & 94.13 & 92.40 & 91.71 & 91.59 & 98.37 & 97.09 & 92.48 & 97.64 & 89.97 & 90.97 & 99.06\\ 
    ViTPose~\cite{vitpose} & ViT-H & 94.73 & 93.18 & 92.06 & 91.75 & 98.92 & 97.77 & 93.42 & 98.63 & 90.49 & 91.62 & 99.47\\
    \bottomrule
    \end{tabular}
    }
    \label{tab:supp_ochuman_mrr}
\end{table}

\begin{table}[tbh]
    \caption{
    The mAP scores of various models on the AP10K-C dataset. Scores are presented from left to right for clean images, all corrupted images, and images corrupted by: Motion Blur, Gaussian Noise, Impulse Noise, Pixelate, JPEG Compression, Color Quantization, Brightness, Darkness, Contrast, and Mask, respectively.
    }
    \centering\scalebox{0.65}{
    \begin{tabular}{rc|cc|cccccccccc}
    \toprule
    \textbf{Method} & \textbf{Backbone} & \textbf{Clean} & \textbf{Corr} & \textbf{Motion} & \textbf{Gauss} & \textbf{Impul} & \textbf{Pixel} & \textbf{JPEG} & \textbf{Quant} & \textbf{Bright} & \textbf{Dark} & \textbf{Contr} & \textbf{Mask} \\
    \midrule\midrule
    SimpleBaseline~\cite{simplebaseline} & Res50 & 68.06 & 49.47 & 46.85 & 41.19 & 41.43 & 63.98 & 60.38 & 52.24 & 60.63 & 41.71 & 29.17 & 57.08\\ 
    SimpleBaseline~\cite{simplebaseline} & Res101 & 68.27 & 50.12 & 47.54 & 43.45 & 43.55 & 64.34 & 60.74 & 53.07 & 60.99 & 41.14 & 29.05 & 57.34\\ 
    ViTPose~\cite{vitpose} & ViT-B & 75.99 & 62.68 & 62.60 & 58.22 & 58.51 & 72.69 & 69.72 & 63.49 & 72.79 & 55.52 & 47.98 & 65.31\\ 
    ViTPose~\cite{vitpose} & ViT-L & 79.60 & 69.04 & 68.74 & 65.01 & 64.87 & 77.01 & 73.95 & 69.15 & 77.62 & 63.23 & 61.55 & 69.31\\
    \bottomrule
    \end{tabular}
    }
    \label{tab:supp_ap10k_map}
\end{table}

\begin{table}[tbh]
    \caption{
    The mAR scores of various models on the AP10K-C dataset. Scores are presented from left to right for clean images, all corrupted images, and images corrupted by: Motion Blur, Gaussian Noise, Impulse Noise, Pixelate, JPEG Compression, Color Quantization, Brightness, Darkness, Contrast, and Mask, respectively.
    }
    \centering\scalebox{0.65}{
    \begin{tabular}{rc|cc|cccccccccc}
    \toprule
    \textbf{Method} & \textbf{Backbone} & \textbf{Clean} & \textbf{Corr} & \textbf{Motion} & \textbf{Gauss} & \textbf{Impul} & \textbf{Pixel} & \textbf{JPEG} & \textbf{Quant} & \textbf{Bright} & \textbf{Dark} & \textbf{Contr} & \textbf{Mask} \\
    \midrule\midrule
    SimpleBaseline~\cite{simplebaseline} & Res50 & 71.77 & 53.51 & 50.72 & 45.45 & 45.81 & 67.68 & 64.42 & 56.16 & 64.93 & 46.31 & 32.74 & 60.91\\ 
    SimpleBaseline~\cite{simplebaseline} & Res101 & 71.92 & 53.99 & 51.38 & 47.59 & 47.74 & 67.96 & 64.61 & 56.92 & 64.96 & 45.15 & 32.33 & 61.29\\ 
    ViTPose~\cite{vitpose} & ViT-B & 79.41 & 66.34 & 66.19 & 62.13 & 62.46 & 76.12 & 73.35 & 67.07 & 76.18 & 59.51 & 51.20 & 69.16\\ 
    ViTPose~\cite{vitpose} & ViT-L & 82.63 & 72.53 & 72.13 & 68.66 & 68.61 & 80.13 & 77.35 & 72.65 & 80.73 & 67.09 & 65.02 & 72.92\\
    \bottomrule
    \end{tabular}
    }
    \label{tab:supp_ap10k_mar}
\end{table}

\begin{table}[tbh]
    \caption{
    The mRR scores of various models on the AP10K-C dataset. Scores are presented from left to right for clean images, all corrupted images, and images corrupted by: Motion Blur, Gaussian Noise, Impulse Noise, Pixelate, JPEG Compression, Color Quantization, Brightness, Darkness, Contrast, and Mask, respectively.
    }
    \centering\scalebox{0.65}{
    \begin{tabular}{rc|cc|ccccccccc}
    \toprule
    \textbf{Method} & \textbf{Backbone} & \textbf{mRR (\%)} &  \textbf{Motion} & \textbf{Gauss} & \textbf{Impul} & \textbf{Pixel} & \textbf{JPEG} & \textbf{Quant} & \textbf{Bright} & \textbf{Dark} & \textbf{Contr} & \textbf{Mask} \\
    \midrule\midrule
    SimpleBaseline~\cite{simplebaseline} & Res50 & 72.68 & 68.83 & 60.53 & 60.87 & 94.01 & 88.71 & 76.76 & 89.08 & 61.29 & 42.86 & 83.87\\ 
    SimpleBaseline~\cite{simplebaseline} & Res101 & 73.42 & 69.64 & 63.65 & 63.80 & 94.24 & 88.97 & 77.73 & 89.34 & 60.26 & 42.56 & 83.99\\ 
    ViTPose~\cite{vitpose} & ViT-B & 82.49 & 82.38 & 76.62 & 77.01 & 95.66 & 91.76 & 83.55 & 95.80 & 73.07 & 63.14 & 85.95\\ 
    ViTPose~\cite{vitpose} & ViT-L & 86.74 & 86.36 & 81.67 & 81.50 & 96.74 & 92.91 & 86.88 & 97.52 & 79.44 & 77.32 & 87.07\\
    \bottomrule
    \end{tabular}
    }
    \label{tab:supp_ap10k_mrr}
\end{table}

We present the detailed results for all models across each corruption type on the three datasets. Results for the COCO-C dataset are shown in Tables~\ref{tab:supp_coco_map},~\ref{tab:supp_coco_mar}, and ~\ref{tab:supp_coco_mrr}. Tables~\ref{tab:supp_ochuman_map},~\ref{tab:supp_ochuman_mar}, and~\ref{tab:supp_ochuman_mrr} show results for the OCHuman-C dataset. Results for the AP10K-C dataset are listed in Tables~\ref{tab:supp_ap10k_map},~\ref{tab:supp_ap10k_mar}, and~\ref{tab:supp_ap10k_mrr}. These results further support the conclusions drawn in Section~\ref{sec:experiment}.

\section{Limitations and Future Work}\label{sec:appendix_disscussion}

While PoseBench offers a comprehensive benchmark for evaluating the robustness of pose estimation methods, it has several limitations:
\begin{itemize}
    \item \textbf{Score of Research Fields.}
    Although we assess 60 pose models, all methods focus solely on 2D pose estimation from monocular images. Evaluations on 3D skeleton-based pose estimation models, multi-view 2D images, and videos remain unexplored.

    \item \textbf{Limited Exploration on Mask Corruptions.} 
    We propose evaluating robustness against occlusions and data loss by randomly occluding a fixed-size rectangle with zero values. Future assessments should consider the occlusion's position, such as covering specific joints.

    \item \textbf{Real-World Scenarios.}
    The applied corruptions are simulated, potentially creating a domain gap between simulated images and real-world corruptions from sensor failures. Collecting and benchmarking real-world corrupted data is essential for a more comprehensive study of robustness.
\end{itemize}

\begin{figure}
    \centering
    \includegraphics[width=\linewidth]{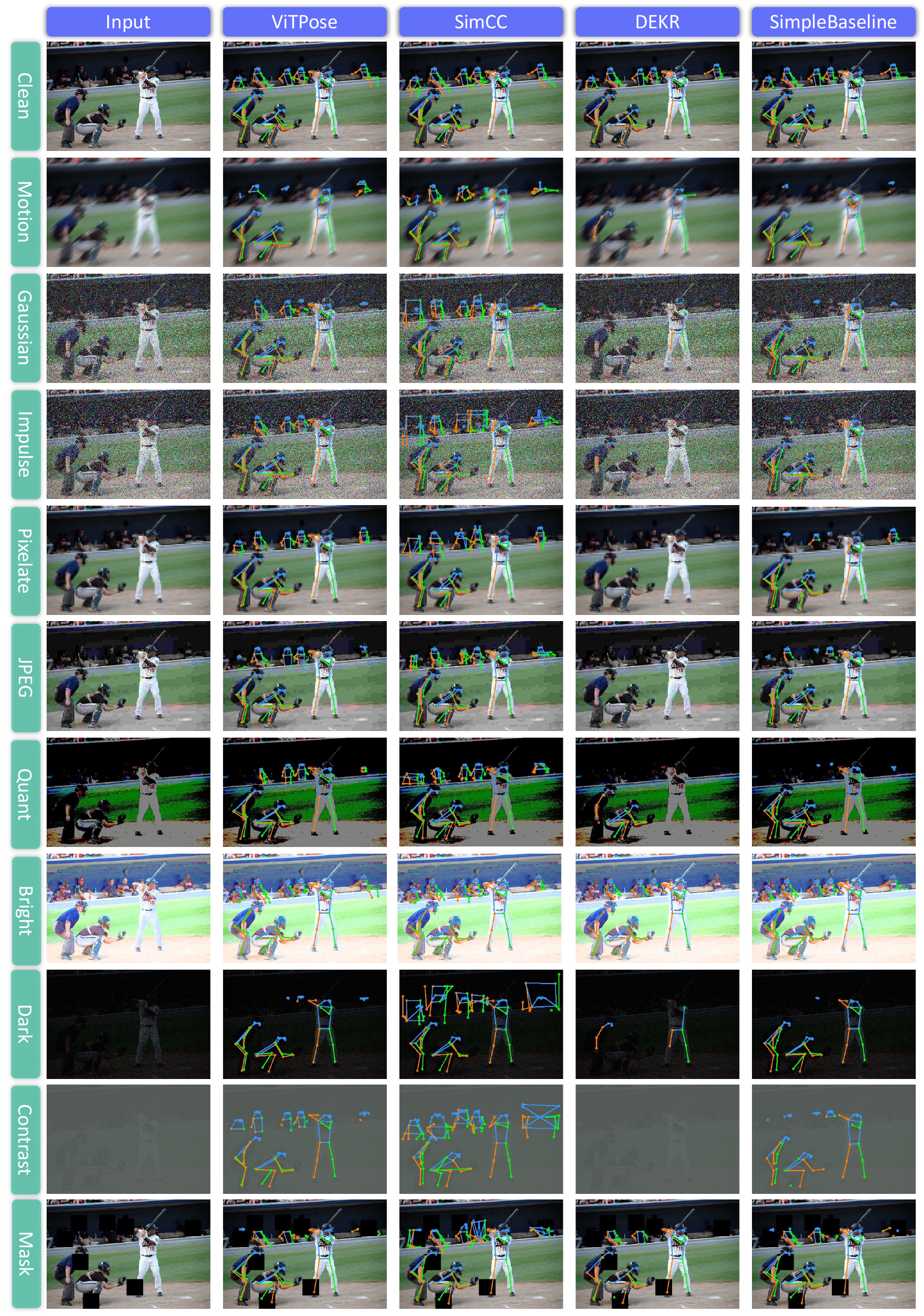}
    \caption{Visualization of human pose estimation results from various methods on the COCO-C dataset under various corruptions.}
    \label{fig:enter-label}
\end{figure}

\end{document}